\documentclass[10pt,twocolumn,letterpaper]{article}

\usepackage{cvpr}
\usepackage{times}
\usepackage{epsfig}
\usepackage{graphicx}
\usepackage{amsmath}
\usepackage{amssymb}
\usepackage{mathtools,empheq}
\usepackage{subfigure}
\usepackage{caption}
\usepackage{verbatim}
\usepackage{xspace}
\usepackage{pbox}

\def\ReduceBeforeCaptionfigspace{\vspace{-0.6cm}}
\def\ReduceAfterCaptionfigspace{\vspace{-0.6cm}}
\newcommand{\Gama}{\boldsymbol{\Gamma}}
\newcommand{\gama}{\boldsymbol{\gamma}}

\usepackage[breaklinks=true,bookmarks=false]{hyperref}

\cvprfinalcopy 

\def\cvprPaperID{1095} 

\begin{document}

\title{The Surfacing of Multiview 3D Drawings via Lofting and Occlusion
Reasoning\\
\texttt{\normalsize Exnhanced version improved over camera-ready}}

\author{Anil Usumezbas\\
  SRI International\\
  201 Washington Rd, Princeton, NJ 08540, USA\\
  {\tt\small anil.usumezbas@sri.com}
  \and
  Ricardo Fabbri\\
  Polytechnic Institute -- Rio de Janeiro State University\\
  R.~Bonfim 25, Vila Amelia, Nova Friburgo RJ 28625-570, Brazil\\
  {\tt\small rfabbri@gmail.com}
  \and
  Benjamin B. Kimia\\
  Brown University\\
  Providence, RI 02912, USA\\
  {\tt\small benjamin\_kimia@brown.edu}
}

\maketitle

\begin{abstract}
  The three-dimensional reconstruction of scenes from multiple views has
  made impressive strides in recent years, chiefly by methods correlating isolated
  feature points, intensities, or curvilinear structure. 
  In the \emph{general setting}, \ie, without requiring controlled acquisition,
  limited number of objects, abundant patterns on objects, or object curves to
  follow particular models, the majority of these methods produce unorganized
  point clouds, meshes, or voxel representations of the reconstructed scene, with
  some exceptions producing 3D drawings as networks of curves.
  Many applications, \eg, robotics, urban planning, industrial design, and hard
  surface modeling, however,
  require structured representations which make explicit 3D curves, surfaces,
  and their spatial relationships. 
  Reconstructing surface representations can now be constrained by the 3D drawing acting
  like a scaffold to hang on the computed representations, leading to increased
  robustness and quality of reconstruction.
  This paper presents one way of completing such 3D drawings with surface
  reconstructions, by exploring occlusion reasoning through lofting algorithms.
\end{abstract}

\section{Introduction}

Dense 3D surface reconstruction is an important problem in computer vision
which remains challenging in general scenarios. Most existing multiview
reconstruction methods suffer from some common problems such as: {\em (i)} Holes in the
3D model corresponding to homogeneous/reflective/transparent image regions, {\em (ii)}
Oversmoothing of semantically-important details such as ridges, {\em (iii)} Lack of semantically meaningful surface features, organization and geometric detail.

\begin{figure}
  \begin{center}
    \includegraphics[width=0.325\linewidth]{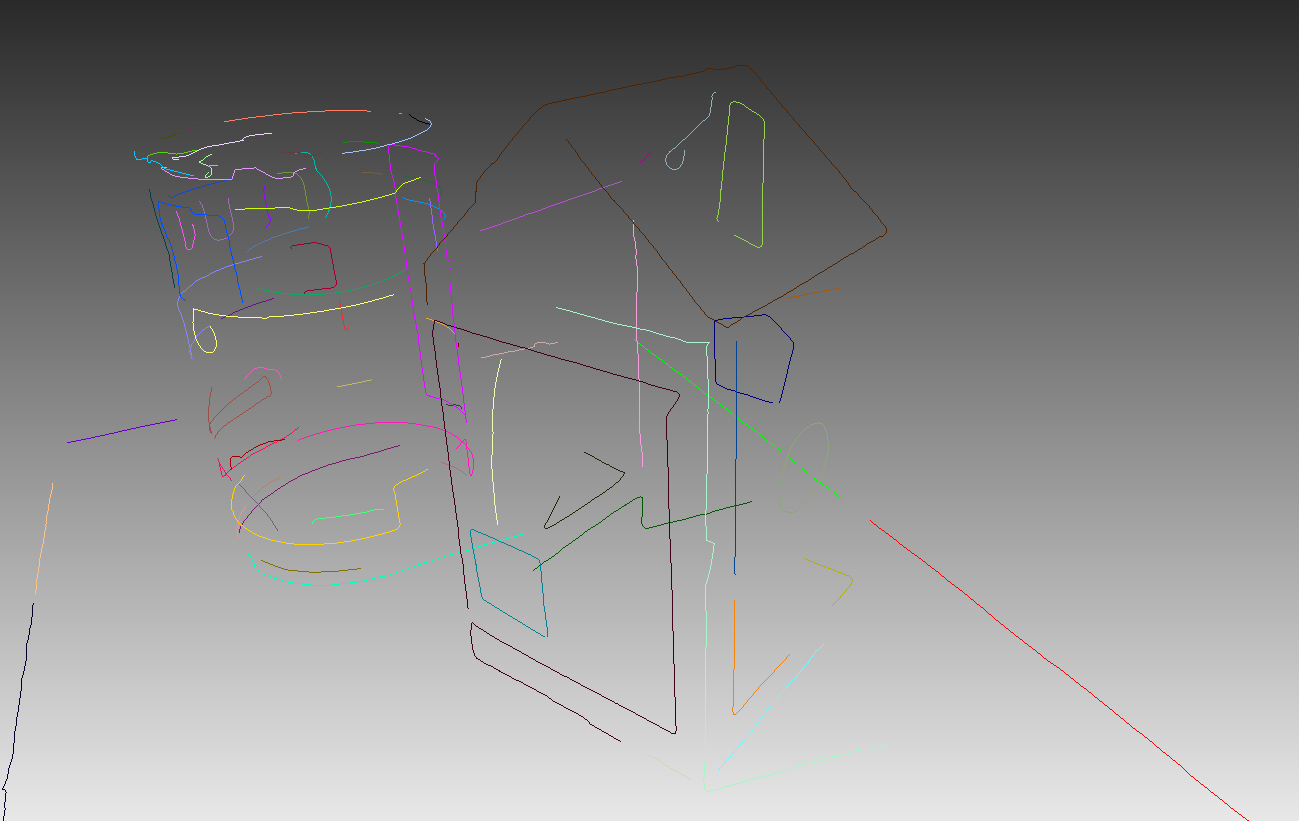}
    \includegraphics[width=0.325\linewidth]{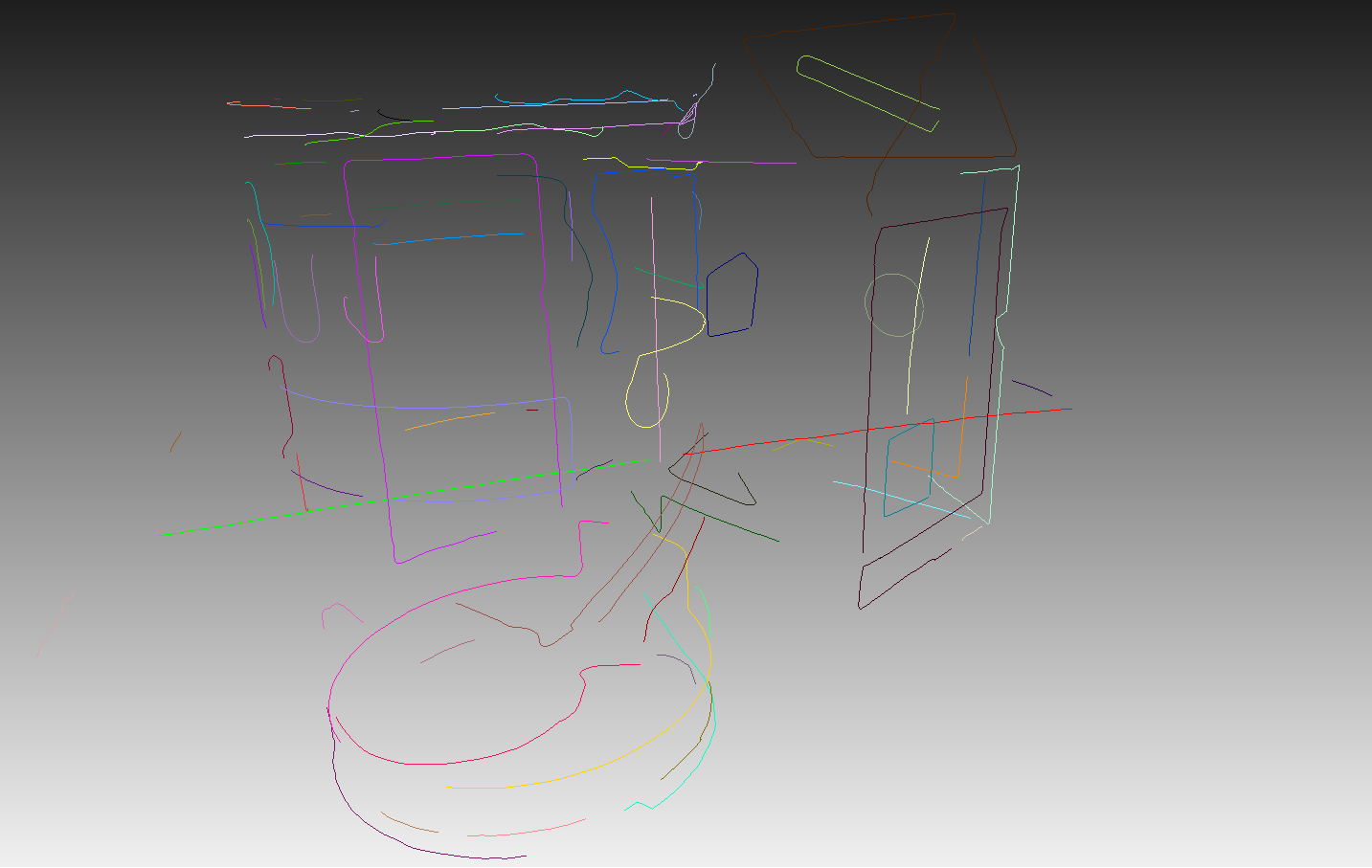}
    \includegraphics[width=0.325\linewidth]{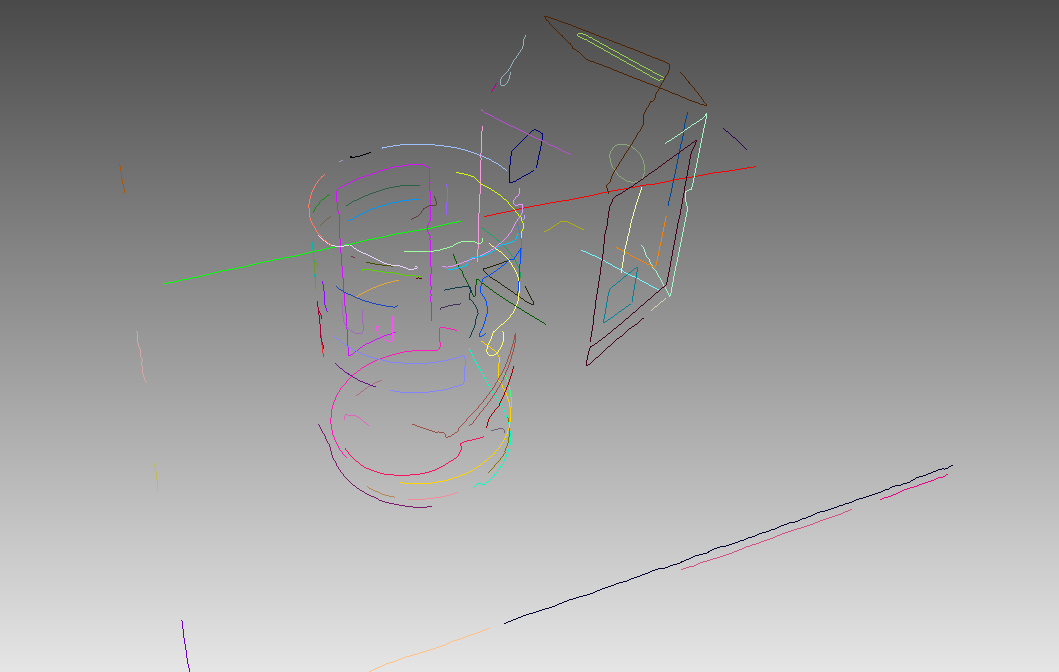}
    \includegraphics[width=0.325\linewidth]{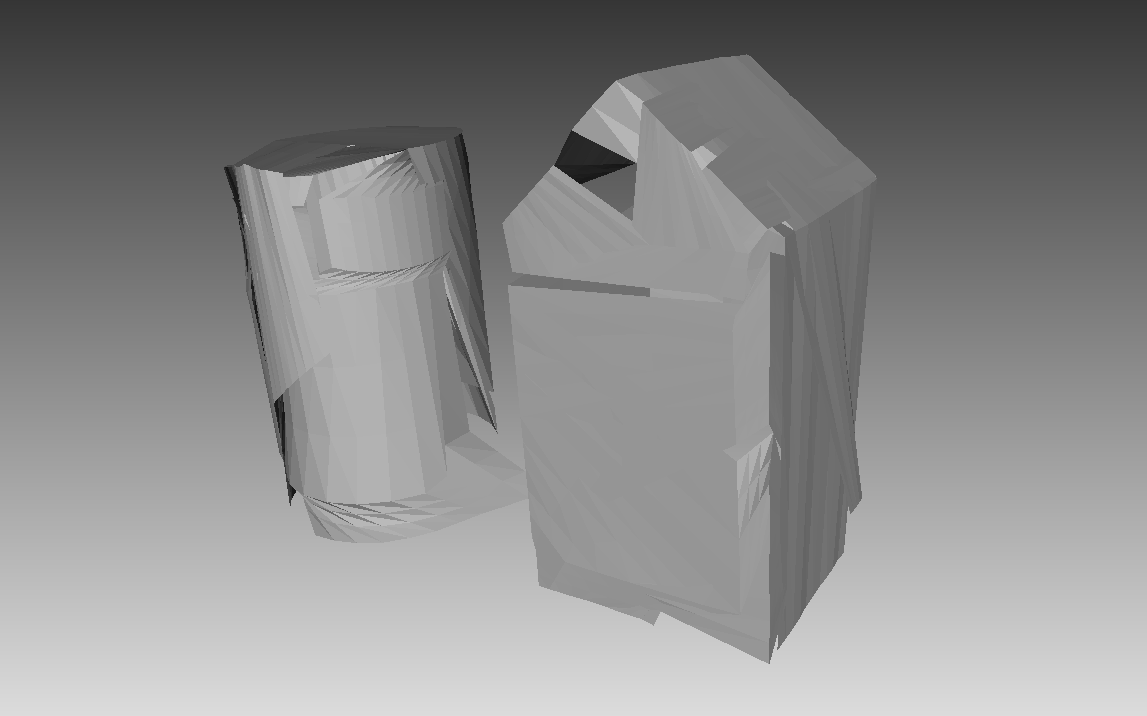}
    \includegraphics[width=0.325\linewidth]{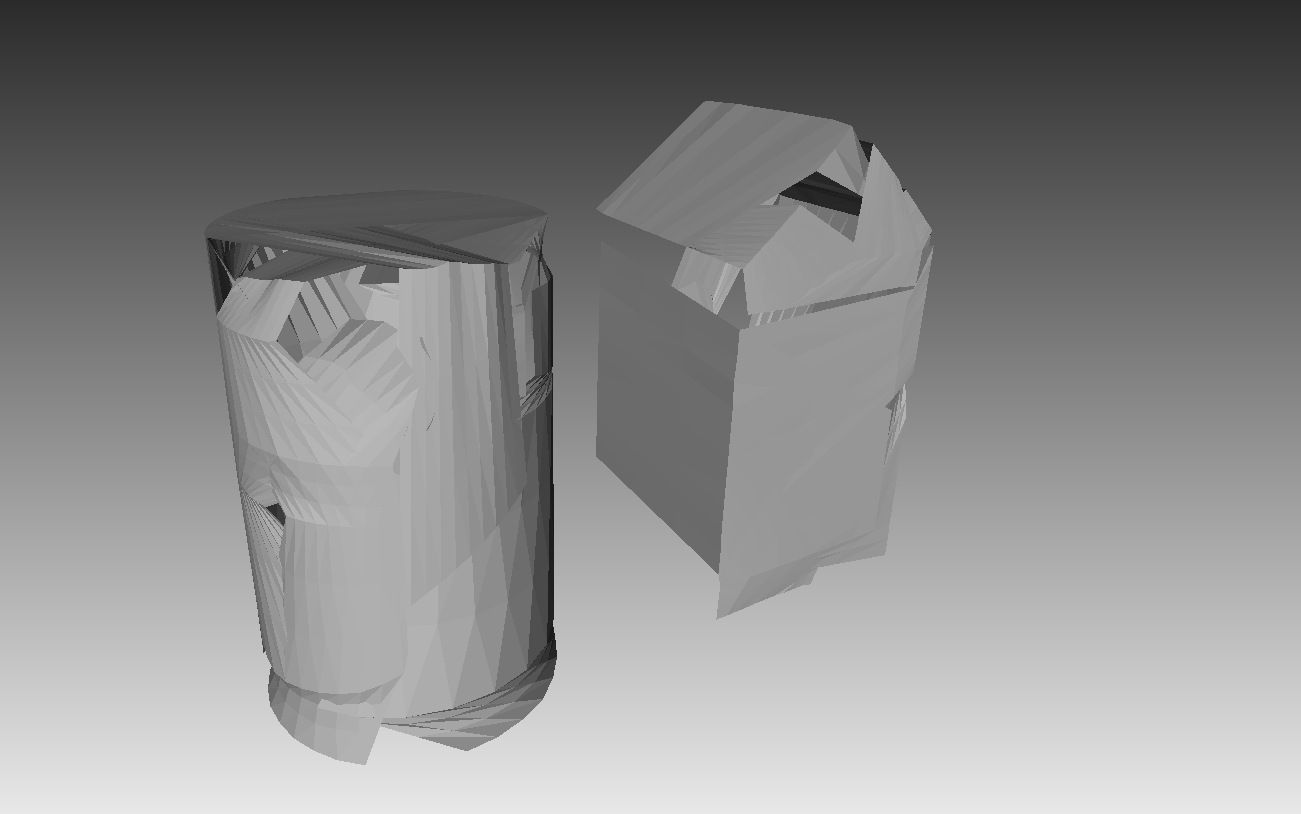}
    \includegraphics[width=0.325\linewidth]{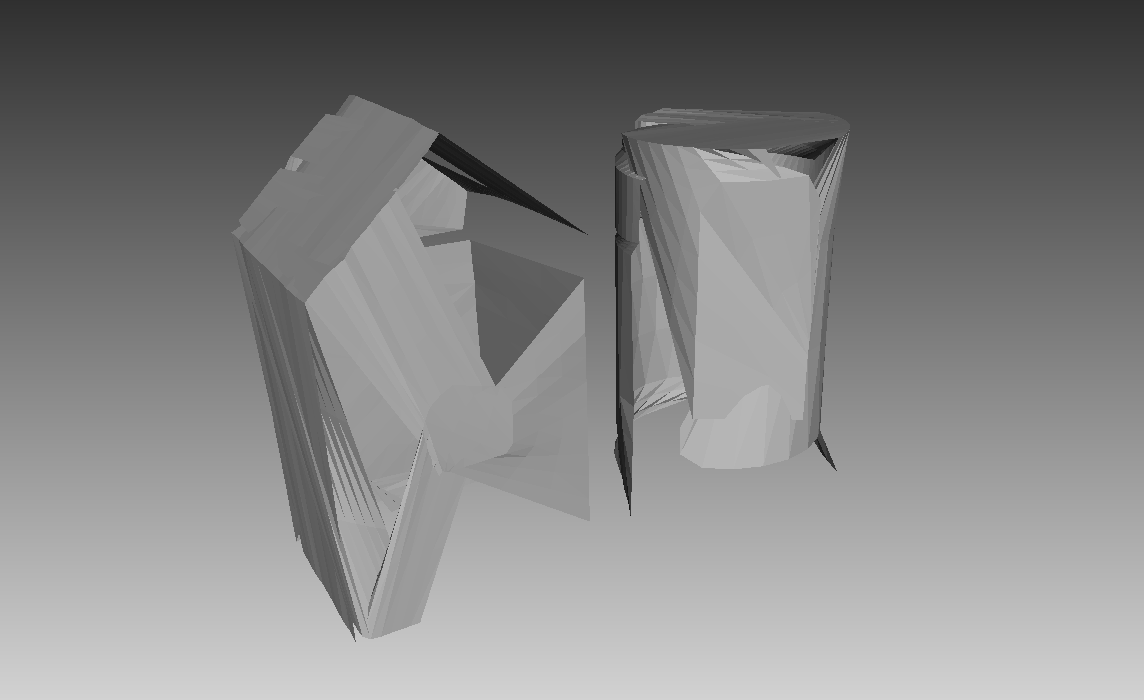}
  \end{center}
  \ReduceBeforeCaptionfigspace
  \caption{
    The proposed approach transforms a 3D curve drawing (top) obtained from a
    fully calibrated set of 27 views, into a collection of dense surface patches
    (bottom) obtained via lofting and occlusion reasoning.}
  \ReduceAfterCaptionfigspace
  \label{fig:recon:lofting:results}
\end{figure}

In computer vision and graphics literature, there has been
scattered but persistent interest in using 3D curves to infer aspects of an
underlying shape~\cite{Maekawa:Ko:GM2002, Zorin:SIGGRAPH2006}, shape-related
features linked to shading~\cite{Bui:etal:CNG2015}, or closed 3D
curves~\cite{Zhuang:etal:ACMTOG2013}. For example, the approach in
Sadri and Singh~\cite{Sadri:Singh:ACMTOG2014} exploits
the \emph{flow complex}, a structure that captures both the topology and the
geometry of a set of 3D curves, to construct an intersection-free triangulated
3D shape. Similarly, the approach in Pan~\etal~\cite{Pan:etal:ACMTOG2015} explores a similar concept with \emph{flow lines}, which are designed to encapsulate principal curvature lines on a
surface. As another example, the approach in Abbasinejad~\etal~\cite{Abbasinejad:etal:SOCG2012} identifies potential surface
patches delineated by a 3D curve network, breaking them into smaller,
planar patches to represent a complex surface. These methods are completely
automated and yield impressive results on a wide range of objects. However, they 
require a complete and accurate input curve network, which is very
difficult to obtain in a bottom-up fashion from image data: there will always be
holes, missed curves, incorrect groupings, noise, outliers, and other
real-world imperfections. Furthermore, these methods are not general, but rather tailored for scenes with objects of relatively clean geometry. Thus, they are not suitable for more
general, large-scale complex scenes that the multiview stereo community
tackles on a regular basis.

We propose a novel and complementary dense 3D reconstruction approach based on
occlusion reasoning and a CAD method called \emph{lofting}, which is the process
of obtaining 3D surfaces through the interpolation of 3D structure curves.
Lofting has primarily been a drafting technique for generating streamlined
objects from curved line drawings that was initially used to design and build
ships and aircrafts. More recently, lofting has become a common technique in
computer graphics and computer-aided design (CAD) applications where a
collection of surface curves are used to define the surface through
interpolation. Even though lofting is a very powerful tool, it does not appear
to be used very much in the multiview geometry applications.  
Employing an existing curve-based reconstruction method, we start with a
calibrated image sequence to build a 3D drawing of the scene in the form of a 3D
graph, where graph links contain curve geometries and graph nodes contain
junctions where curve endpoints meet.  We propose to use the 3D drawing of a
scene as a scaffold on which dense surface patches can be placed on, see
Figure~\ref{fig:recon:lofting:results}.
Our approach relies on the availability of a ``3D drawing'' of the surface, a
graph of 3D curve fragments reconstructed from calibrated multiview observations
of an object \cite{Usumezbas:Fabbri:Kimia:ECCV16}. Observe that such a 3D
drawing acts as a scaffold for the surface of the object in that the drawing
breaks the object surfaces into 3D surface patches, which are glued on and
supported by the 3D drawing scaffold. Our approach then is based on selecting
some 3D curve fragments from the 3D drawing, forming surface hypotheses from
these curve fragments, and using occlusion reasoning to discard inconsistent
hypotheses.

Aside from yielding a useful and semantically-meaningful intermediate
representation, reconstructing surfaces by going through curved structures
closely replicates the human act of drawing: As in a progressive drawing, the
basis is independent of illumination conditions and other details. For instance,
photometry/shading/reflectance can be incorporated later on either as hatchings
or progressively refined as fine shading; multiple renderings can be performed
from the same basis. Even challenging materials such as the ocean surface can be
rendered on top of a curve basis. This approach also has the advantage of
scalability, since it allows for a very large 3D scene to be selectively and
progressively reconstructed.

This paper is organized as follows: Section~\ref{sec:curve_drawing} reviews the
state-of-the-art in generating a 3D drawing of a scene observed under calibrated
views. Section~\ref{sec:lofting} reviews lofting and describes how a surface is
generated from a few curve fragments lying on the surface.
Section~\ref{sec:auto-lofting} describes how 3D surface patch hypotheses are
generated from a 3D drawing, and how occlusion consistency is used to take out
non-veridical hypotheses. Section~\ref{sec:lofting-results} deals with several
technical challenges, which require a regularization of the 3D drawing so that
surface patches can be robustly inferred. Section~\ref{sec:lofting-results}
presents experimental results, a comparison with
PMVS~\cite{Furukawa:Ponce:CVPR2007,Furukawa:Ponce:PAMI2010}, and quantification
of reconstruction accuracy.

\section{From Image Curves to a 3D Curve Drawing}
\label{sec:curve_drawing}

Our multiview stereo method is based on the idea of using 3D curvilinear
structures as boundary conditions to hypothesize the simplest 3D surfaces that
would be explained by these boundaries. The 3D curvilinear structure that is
needed is obtained by correlating image curves in calibrated multiview imagery
to reconstruct 3D curve fragments, which are organized as a graph and referred
to as ``3D Curve Drawing'' \cite{Usumezbas:Fabbri:Kimia:ECCV16}. Since this
paper requires a 3D curve drawing available, we summarize the work of
\cite{Usumezbas:Fabbri:Kimia:ECCV16} on which we rely.

The 3D curve drawing is built on a series of steps. First, the image is
pre-processed to obtain edges using robust, third-order operators which give
highly-accurate edge information \cite{Tamrakar:Kimia:ICCV07}. Second, a
geometric linker groups edges into curves \cite{Yuliang:etal:CVPR14} which
claims to improve on grouping errors and extent of outliers. This results in
image curve fragments ${\gama_i^v, i=1,\dots,M^v}$ for each view $v=1,\dots,N$.
Third, pairs of curves $(\gama_{i_1}^{v_1}, \gama_{i_2}^{v_2})$ from two
``hypothesis views'' $v_1$ and $v_2$, which have significant epipolar overlap,
are used to generate putative candidate reconstructions $\Gama_k, k=1,\dots,K$.
These candidate reconstructed curves are gauged against image evidence on other
projected views called ``confirmation views'' and if there is sufficient support
for a 3D curve candidate, it is confirmed and otherwise rejected. This results
in a set of unorganized 3D curve fragments called the ``3D Curve Sketch''.

This representation indeed resembles a sketch. 3D curve fragments in this sketch
are often redundant since they came from multiple hypotheses, are often
overfragmented due to partial epipolar overlap, feature a nontrivial level of
clutter, and most importantly, are unorganized in that the topological
relationship of 3D curve fragments is not available. The recent work of
\cite{Usumezbas:Fabbri:Kimia:ECCV16} deals with these issues, and constructs a
graph of 3D curve fragments referred to as a 3D drawing of the scene.

Our approach requires 3D curve fragments and their topological relationships. To
the best of our knowledge, the approach in \cite{Usumezbas:Fabbri:Kimia:ECCV16}
is the state of the art in curve-based multiview stereo. However, any other
method that can give 3D curve fragments organized in a topological graph can be
used by our approach as well.

\section{Bringing Lofting Into Multiview Stereo}\label{sec:lofting}

\begin{figure}
  \begin{center}
    \includegraphics[width=0.49\linewidth]{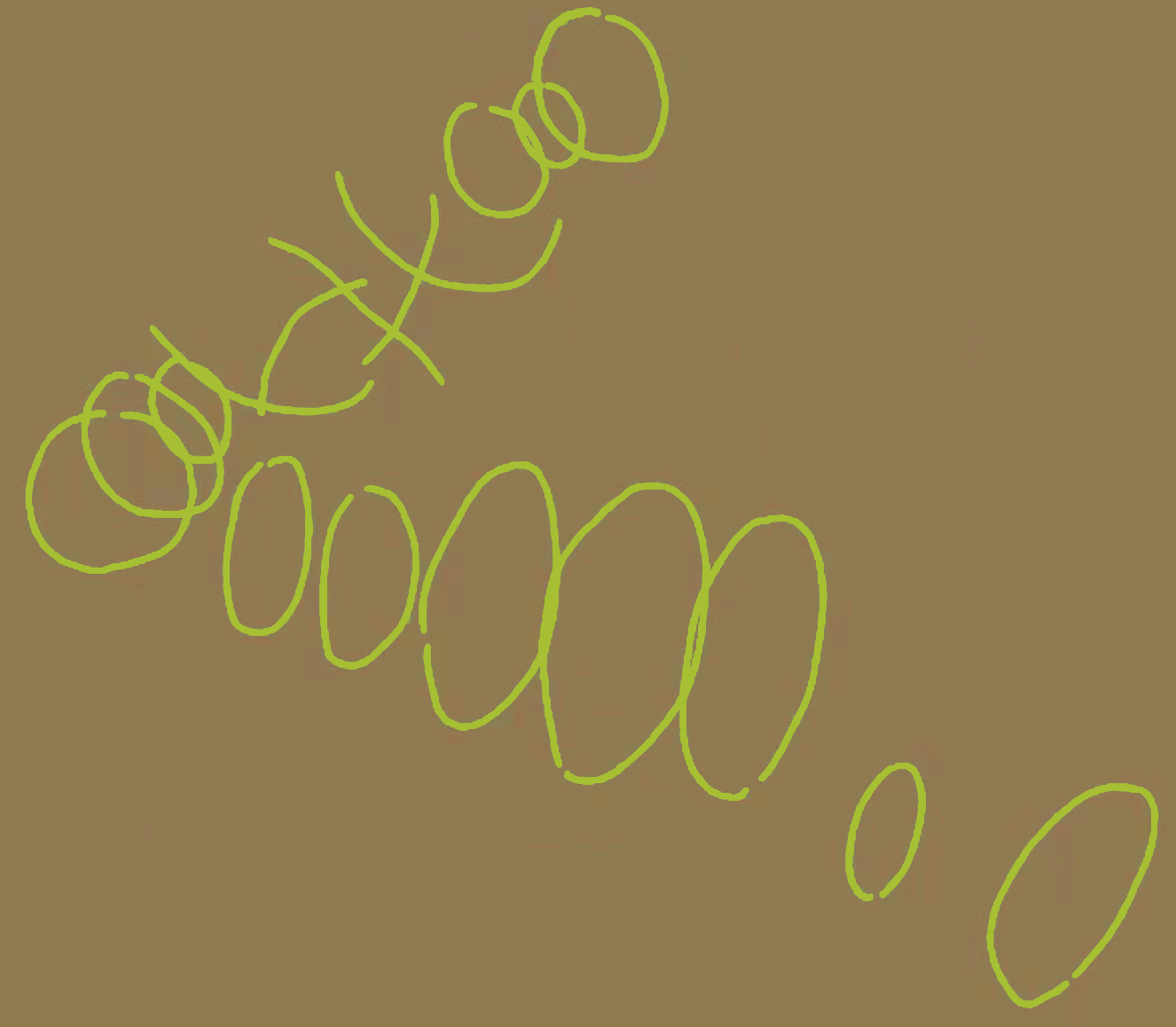}
    \includegraphics[width=0.49\linewidth]{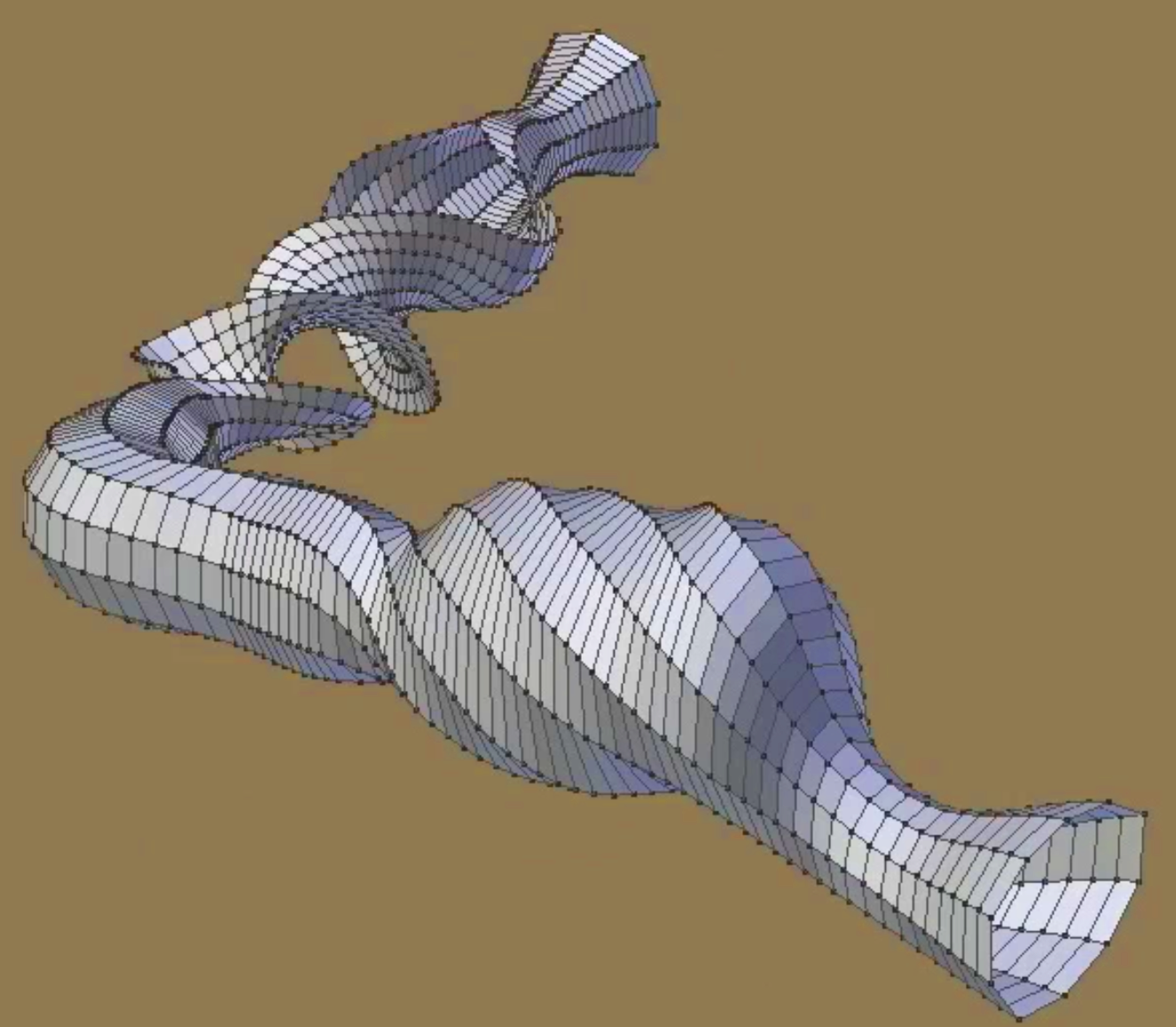}
  \end{center}
  \ReduceBeforeCaptionfigspace
  \caption{From open and closed curves (left), lofting produces smooth surfaces
  (right).}
  \ReduceAfterCaptionfigspace
  \label{fig:lofting}
\end{figure}

Lofting is graphics technique for shape inference from a set of 3D curves, a
term with roots in shipbuilding to describe the molding of a hull from
curves~\cite{Bole:STR2015}.  Designers often use such intermediate, curve-based
representations (sketches, graphs, drawings) to outline 3D shape, as they
compactly capture rich 3D information and are easy to customize.  Through
lofting, these 3D curves are used to interactively model smooth surfaces,
Figure~\ref{fig:lofting}.
Implementations of lofting are commonplace in interactive CAD~\cite{Blender,
Wu:etal:1977, Nealen:etal:SIGGRAPH2007, Morigi:Rucci:SBIM2011,
Grimm:Joshi:SBIM2012, Abbasinejad:etal:SOCG2012, Das:etal:SBIM2005,
Nam:Chai:CNG2012}, and
applications~\cite{Lin:etal:CII1997,Beccari:etal:CAD2010,Tustison:etal:CVPR2004}.
Lofting has not yet spread to 3D computer vision, where fully-automated
image-based modeling is the norm.  This work leverages lofting to build a
fully-automated, dense multiview stereo reconstruction pipeline.

Given 3D curves $\Gama_1, \Gama_2, \dots \Gama_n$ forming the partial boundary
of a surface, lofting produces a smooth surface passing through them which is
sought to be `simple': smooth, avoiding holes and degeneracies such as
self-intersections.  Earlier approches formulated this as surface deformation
with parameters estimated to fit the prior into a 3D curve
outline~\cite{Chiyokura:Fumihiko:SIGGRAPH1983, Kraevoy:etal:SBIM2009}.
Approaches using functional optimization~\cite{Morigi:Rucci:SBIM2011,
Nealen:etal:SIGGRAPH2007,Sorkine:Cohenor:SMI2004,Welch:Witkin:SIGGRAPH1994,Bobenko:Schroder:SGP2005,Moreton:Sequin:SIGGRAPH1992}
employ generic objectives, such as least squares and integral of squared
principal curvatures, and the result depends on this choice, leading to
overfitting or oversmoothing. These approaches cannot easily handle complex
shapes with many self occlusions~\cite{Lin:etal:CII1997}.
Other algotihms include those based on B-splines~\cite{Woodward:CAD1988,Park:etal:IJAMT2004}.

\begin{figure}
  \begin{center}
    \includegraphics[width=0.8\linewidth]{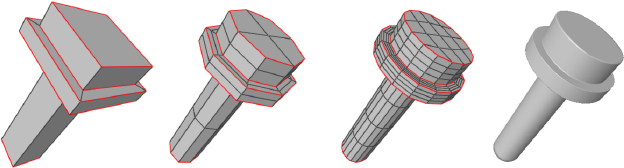}
  \end{center}
  \ReduceBeforeCaptionfigspace
  \caption{Application of subdivision resulting in
    a high-poly surface (manually marked hard edges in
    red)~\cite{Lavoue:etal:techreport}.
    \ReduceAfterCaptionfigspace
  } \label{fig:subdivision}
\end{figure}

We have chosen lofting based on subdivision surfaces, a well-known graphics
technique that divides the faces of a coarse input mesh via a recursive sequence
of transforms or subdivision schemes, yielding smooth high-poly meshes,
Fig.~\ref{fig:subdivision}.  Subdivision is widely used in a number of graphics
problems~\cite{Peters:Reif:ACMTOG1997,DeRose:etal:SIGGRAPH1998}, such as surface
fitting~\cite{Suzuki:etal:CGA1999,Takeuchi:etal:2000,Ma:etal:CAD2004},
reconstruction~\cite{Hoppe:etal:SIGGRAPH1994, Maekawa:Ko:GM2002,
Zorin:SIGGRAPH2006}, and lofting itself~\cite{Nasri:CAGD1997, Nasri:CAGD2000,
Nasri:CGF2003,Nasri:etal:SMI2001,Nasri:Abbas:GMP2002,Catalano:subdivision:book}.
Combined subdivision schemes~\cite{Levin:SIGGRAPH1999,Levin:CAGD1999} translate
conditions on the limit surface to conditions on the scheme itself, and allow
subdivision to be adjusted near the curve network and boundary conditions beyond
subdivision or spline curves. Subdivision surfaces provide a simple standard
framework, with more powerful schemes compared to other techniques; meshes with
complex constraints at corners can be handled with greater
ease~\cite{Schaefer:etal:SGP2004}.
We leverage~\cite{Schaefer:etal:SGP2004}, which takes open 3D polygonal lines
terminating in a set of corners -- as in our 3D drawing, but interactively
generated. We have augmented it to automatically reorganize the curve network
prior to lofting, and with additional heuristics to avoid degeneracies.  The
result is a lofting approach that can: i) take any number of boundary curves
partially or completely covering the boundary of the desired surface, and ii)
handles topological inconsistencies, self-intersections, discontinuities and
other geometric artifacts.  A brief description of our lofting stage follows.

\begin{figure}
  \begin{center}
    \includegraphics[height=1.3cm]{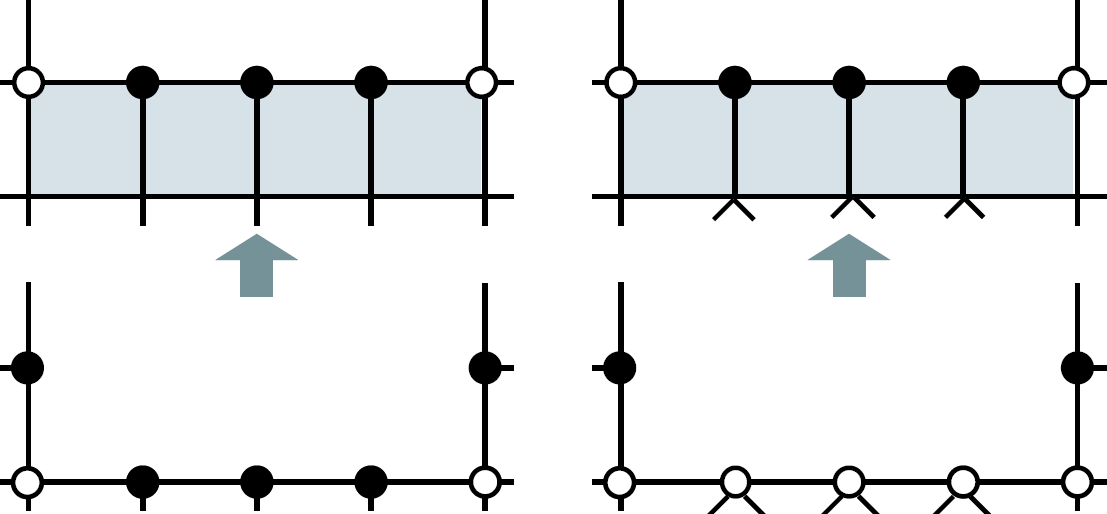}
    \includegraphics[height=1.3cm]{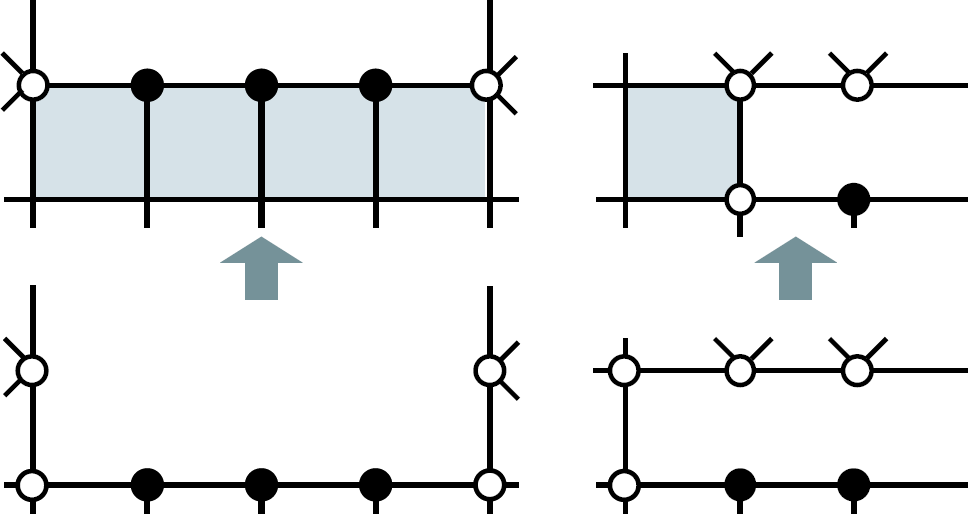}
  \end{center}
  \ReduceBeforeCaptionfigspace
  \caption{Quaqdrangulation in lofting; depending on the configuration of
    special interior vertices on a chain, one of these
    edits are applied to obtain a base
    mesh topology~\cite{Schaefer:etal:SGP2004}.}
  \ReduceAfterCaptionfigspace
  \label{fig:chain:advance}
\end{figure}

\noindent\textbf{Skinning:} quadrangulates the input curves to construct a quad
topology base mesh without the final
geometry~\cite{Schaefer:etal:SGP2004,Piegl:Tiller:CAD1996,
Kaklis:Ginnis:CAGD1996, Nasri:etal:CGA2003}.  Skinning does not produce accurate
shape approximation, but mainly avoids vertices lacking curvature
continuity~\cite{Loop:SGP2004}.  Given a closed 3D curve $\Gama =
(s_1,\dots,s_n)$, a chain is a subsequence $\Gama_i^{i+k} =
(s_i,\dots,s_{i+k})$, $i = 1, \dots, n+k$. The topology of the base mesh
$\lambda$ is constructed by a sequence of chain advances on $\Gama$: given
$\Gama_i^{i+k}$, this adds a layer of $k$ quads to $\lambda$ bounded below by
$\Gama_i^{i+k}$ and above by a new chain $\bar \Gama_j^{j+k} =
(s_j,\dots,s_{j+k})$ on the interior of the resulting patch $\lambda$. $\Gama$
is replaced by $\tilde \Gama = \Gama_1^{i-1} \cup \bar \Gama_j^k \cup
\Gama_{i+k+1}^n$. 
Depending on the configuration of special interior vertices, different types of
advances apply~\cite{Schaefer:etal:SGP2004}, Fig.~\ref{fig:chain:advance}.

\textbf{Fairing} computes the positions of the vertices in $\lambda$ by
minimizing ``fairness'' energy, a thin-plate
functional~\cite{Schaefer:etal:SGP2004}.
\textbf{Subdivision} is then applied with a modified version of Catmull-Clark
schemes~\cite{Schaefer:etal:SGP2004}, yielding a fine mesh, see
Figure~\ref{fig:subdivision}.


\section{Automated Multiview Reconstruction Using Lofting}
\label{sec:auto-lofting}

%

In the previous two sections, we described: {\em (i)} The concept of a {\em 3D
curve drawing}, a graph of 3D contour fragments and a method for deriving it
from a set of calibrated multiview imagery, and {\em (ii)} the concept of
lofting which reconstructs 3D surface meshes bounded by a set of given contour
fragments. We now describe how {\em pairs of curve fragments} selected from the
3D curve drawing give rise to 3D surface hypotheses. These hypotheses are then
ruled out when they predict occlusions which are not consistent with the input
data. The remaining hypotheses yield a set of occlusion-consistent surface
patches. In the following, we first describe the process of hypothesis formation
and then testing of formed hypotheses for occlusion consistency.

\begin{figure}
  \begin{center}
    \includegraphics[width=0.45\linewidth]{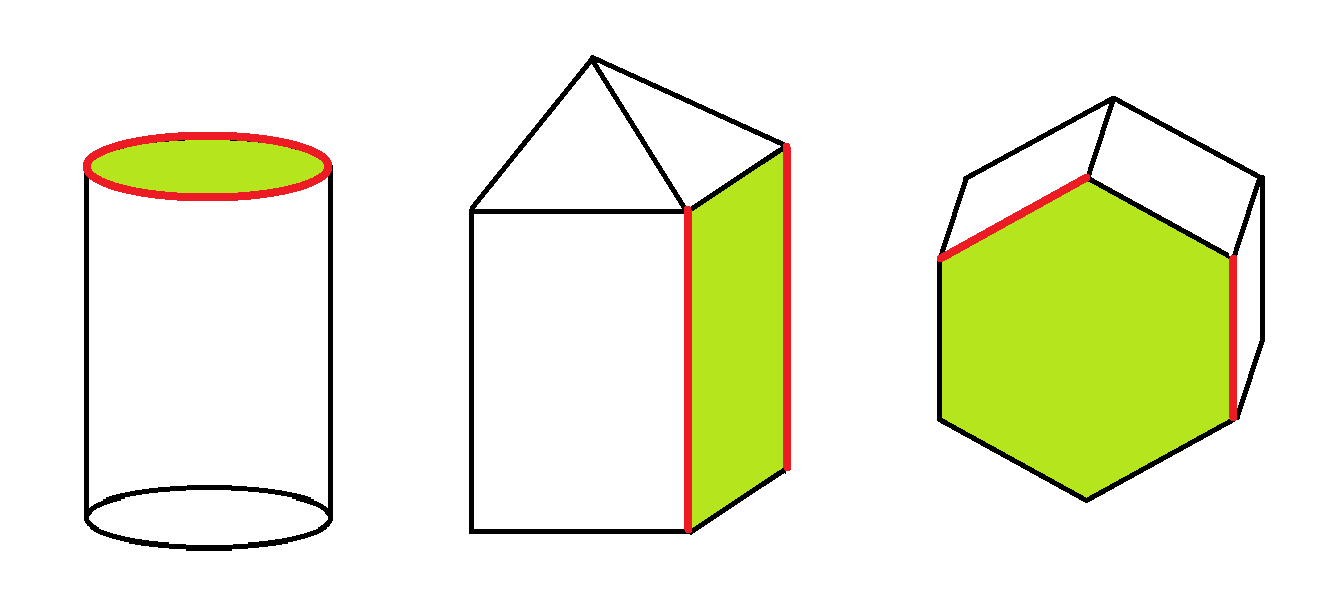}
  \end{center}
  \ReduceBeforeCaptionfigspace
  \caption{A schematic of a simple shapes where a surface patch (green) is
  represented by a pair of curves (red); in the case of closed curves, a pair is
  not necessary.
  }
  \ReduceAfterCaptionfigspace
  \label{fig:shape:schematic}
\end{figure}

\noindent{\textbf{Forming Surface Patch Hypotheses:} Ideally, any subset of
curve fragments should be able to form surface hypotheses, but this is clearly
intractable; even if curve fragments are long, noiseless and salient (a critical
factor as we shall see in Section~\ref{sec:reorg}), they number in the order of
100 curves or so. Note that surface patches that arise from closed curves are a
special case and these be identified and processed a priori. The remaining
surface patches involve at least two curve fragments but typically more, say
around 3-5. Then, pairs of curve fragments can be used as entry level
hypotheses, Figure~\ref{fig:shape:schematic}.

The pool of curve fragments from which pairs are selected is restricted to those
with a minimal length constraint, $L > \tau_{length}$. This threshold is learned
from data and is typically around a few centimeters for our data. The distance
between two 3D curves is defined as the average point-to-curve distance for all
the samples on both curves. The typical 3D curve proximity threshold
$\tau_{\alpha}$, which is also learned from data, is around 15-20 cm. 


Third, in addition to length and pair proximity, curvature of the reconstructed
surface is a cue to whether it is veridical. This is because object surfaces are
typically not as convoluted as surfaces arising from unrelated cues. We use
average Gaussian curvature, \ie Gaussian curvature at every point on the surface
averaged over all surface points, and a threshold $\tau_G$ which is also
learned. It should be noted that every curve pair generates two surface
hypotheses: each endpoint in a given curve can pair with two possible endpoints
on the other curve in the pair. The surface hypotheses with lower average
Gaussian curvature is the one that is selected, if it is above $\tau_G$,
Figure~\ref{fig:loft:options}. See Figure~\ref{fig:lofting:samples} for a
collection of sample surface hypotheses obtained this way.

\begin{figure}
  \begin{center}
    \includegraphics[width=0.485\linewidth]{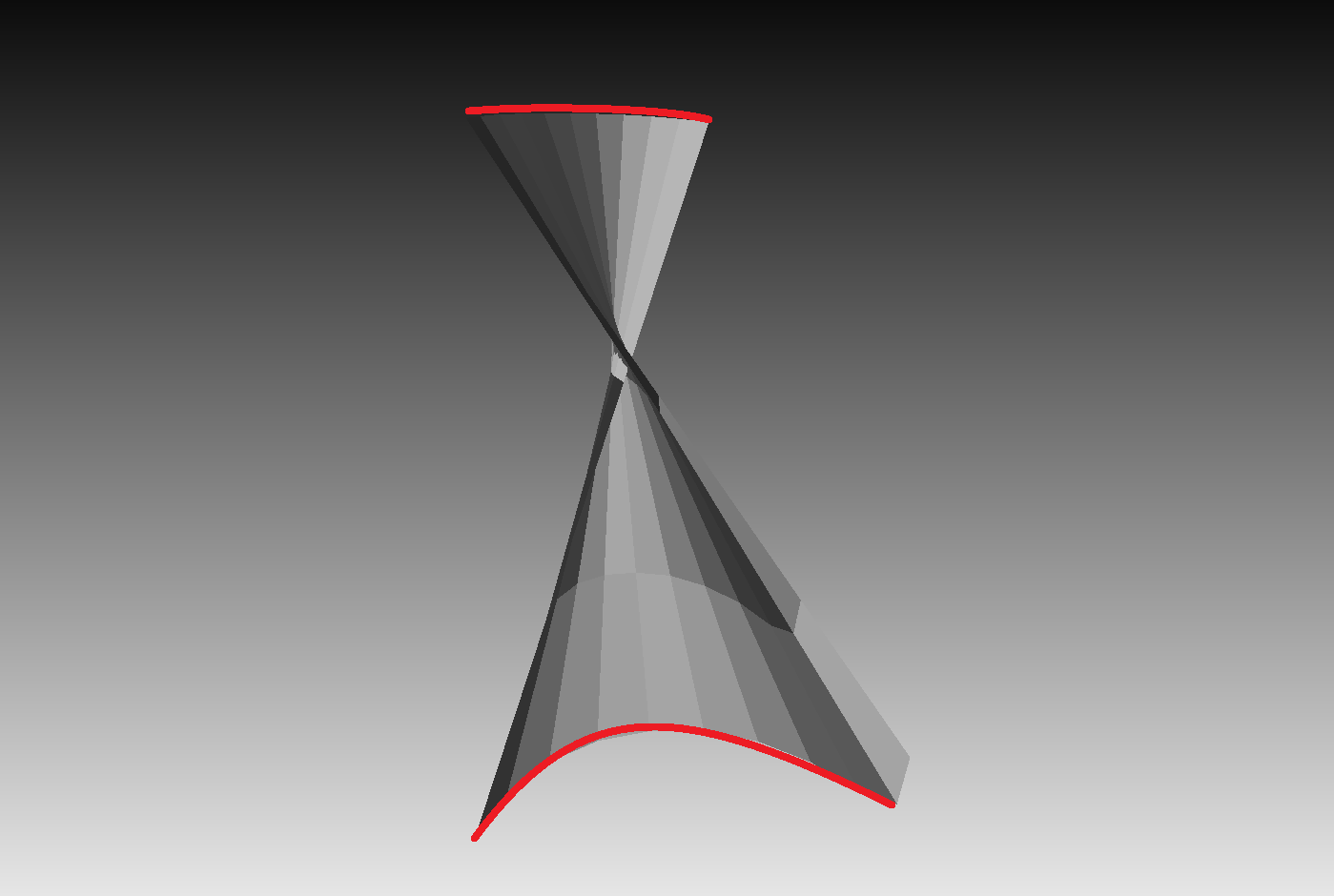}
    \includegraphics[width=0.485\linewidth]{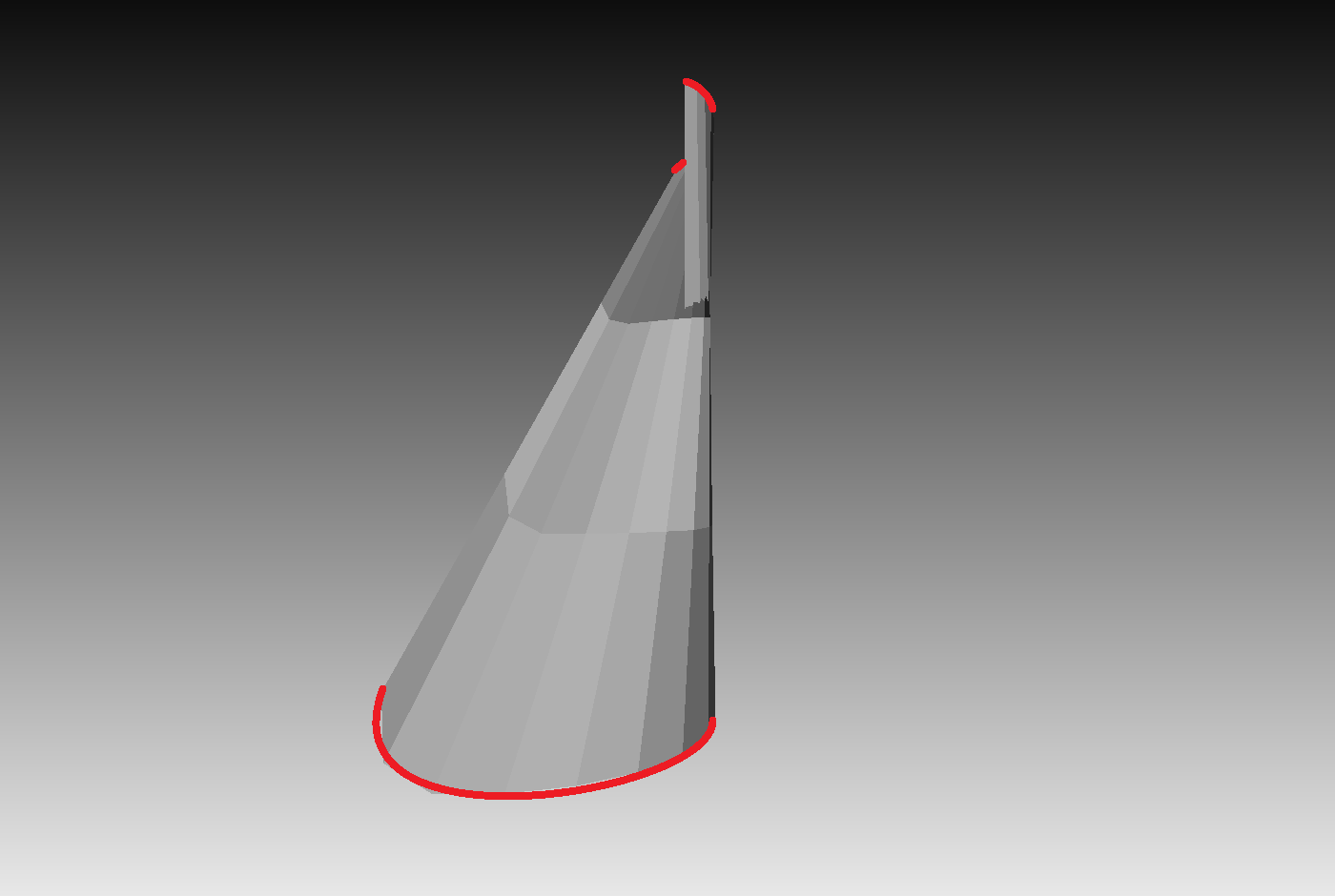}
    \includegraphics[width=0.485\linewidth]{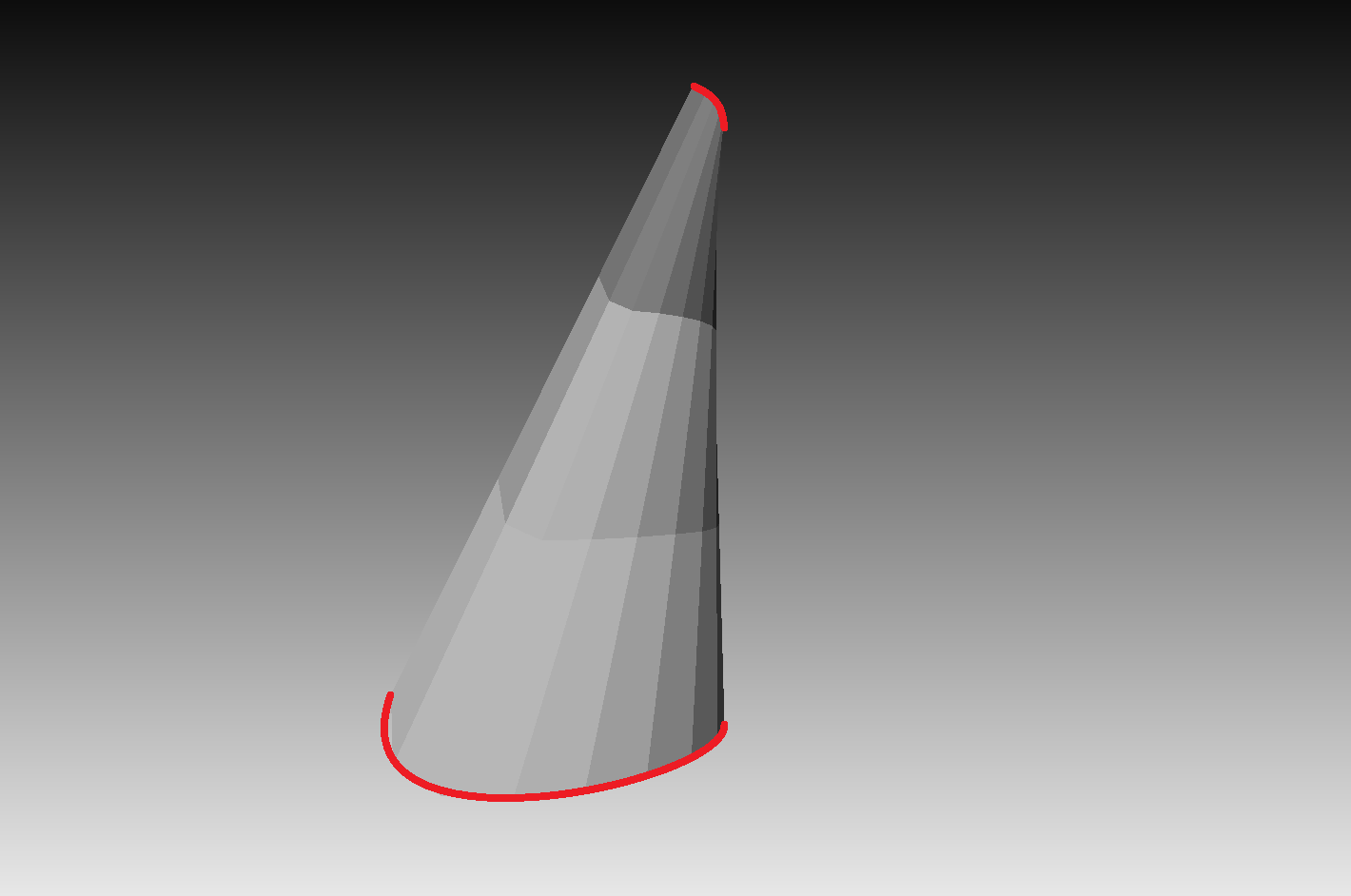}
    \includegraphics[width=0.485\linewidth]{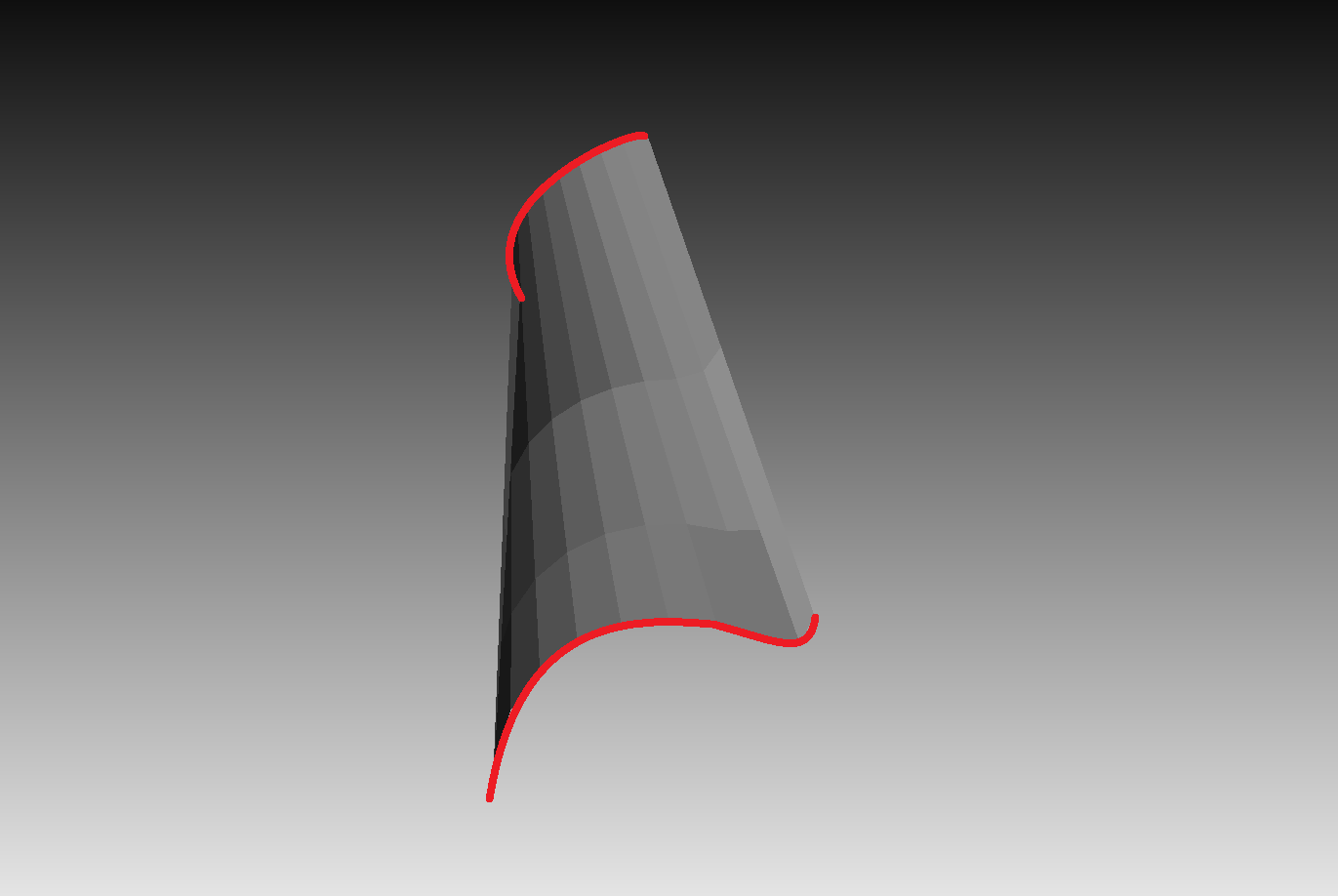}
  \end{center}
  \ReduceBeforeCaptionfigspace
  \caption{There is an inherent ambiguity in reconstructing a surface from two
    curve fragments arising from which endpoints are paired (top row vs. bottom
    row). 
    When two curve fragments do belong to a veridical surface, one of the two
    reconstructions generally has much lower average Gaussian curvature than the
    other and this is a cue as to which one is veridical. When the pairing of
    curve fragments is incorrect in that no surface exists between them, both
    reconstructs have high average Gaussian curvature, a cue to remove outliers.
  }
  \label{fig:loft:options}
\end{figure}

\begin{figure}
  \begin{center}
    \includegraphics[width=0.325\linewidth]{figs/loft-option-3.png}
    \includegraphics[width=0.325\linewidth]{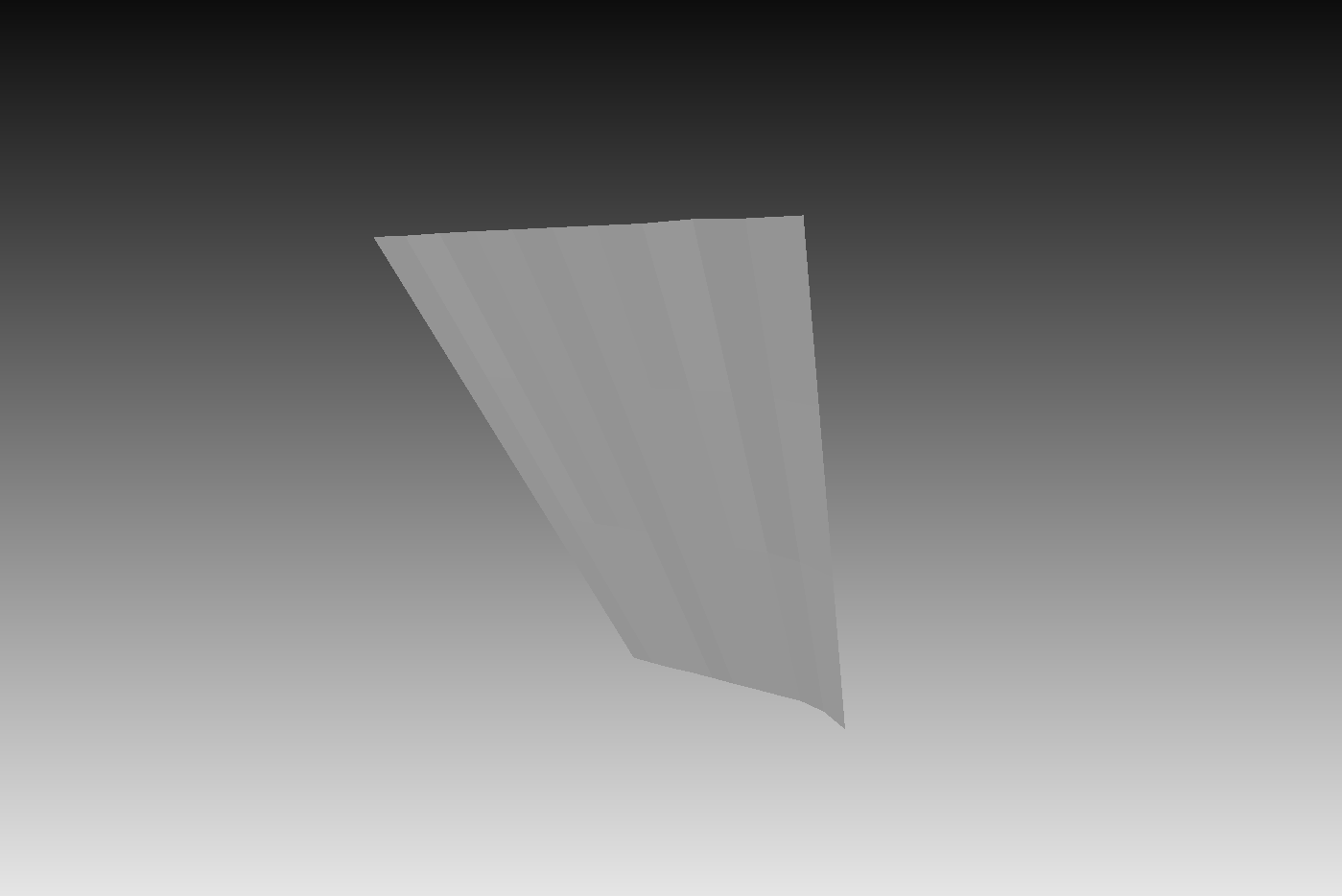}
    \includegraphics[width=0.325\linewidth]{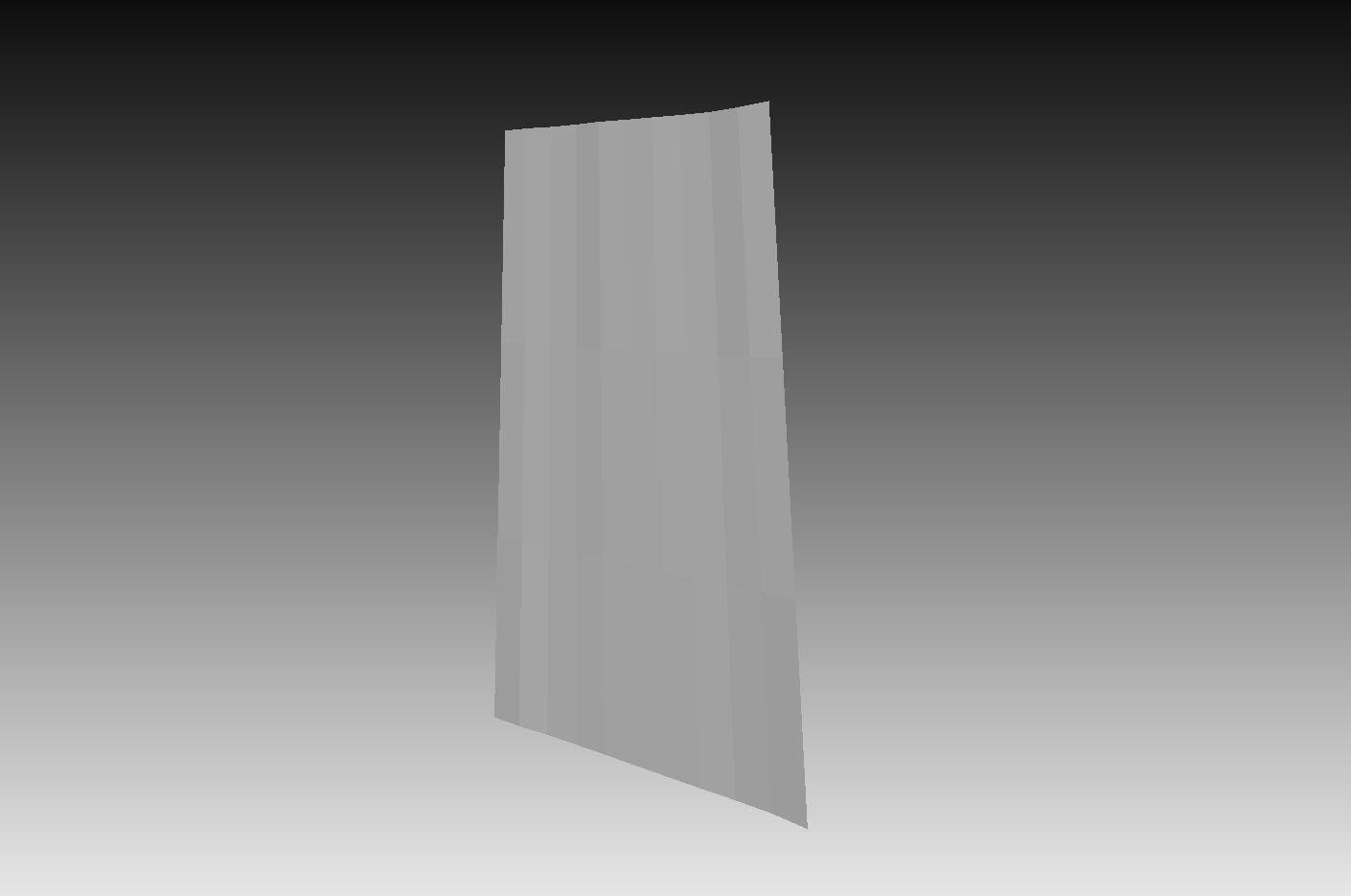}
    \includegraphics[width=0.325\linewidth]{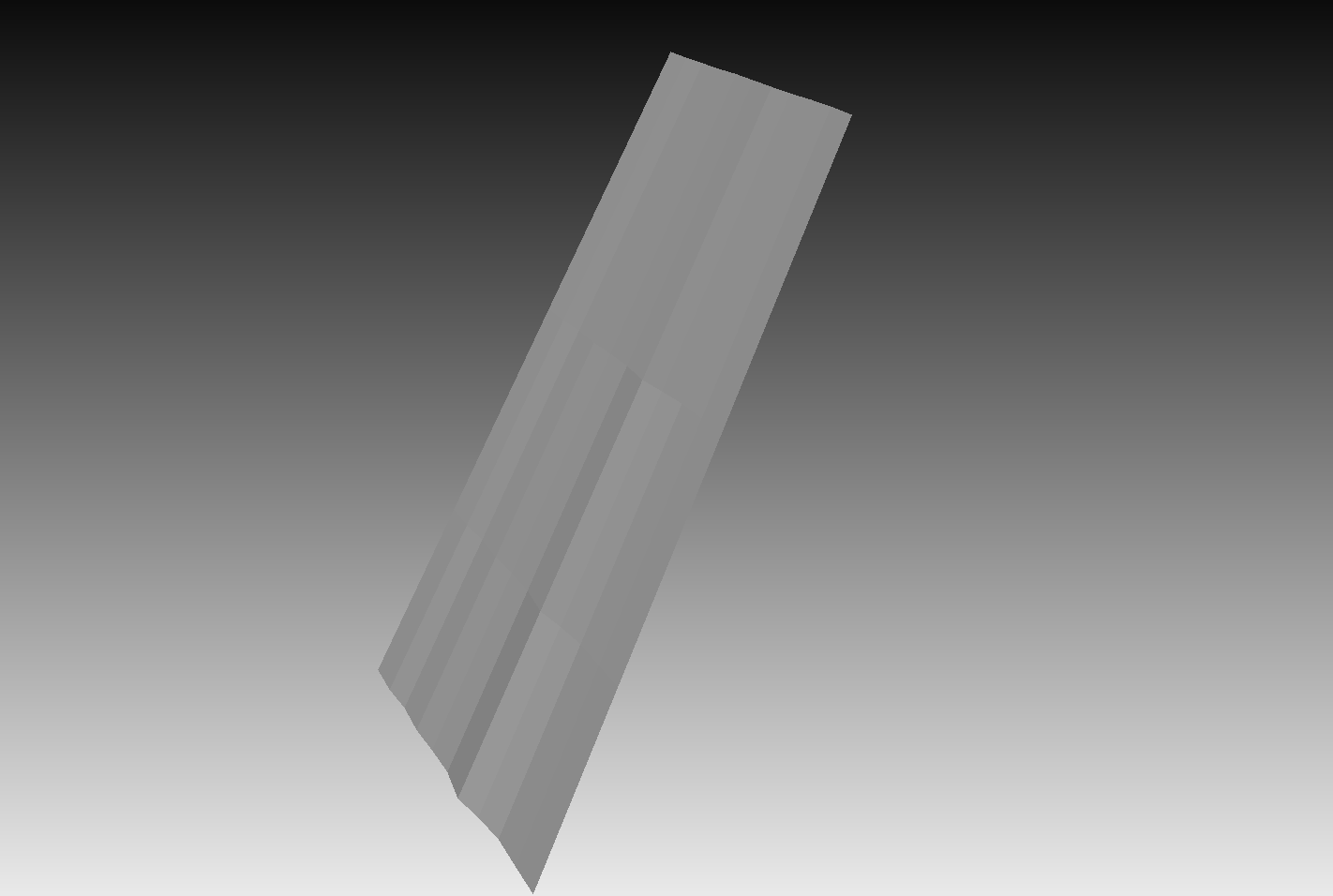}
    \includegraphics[width=0.325\linewidth]{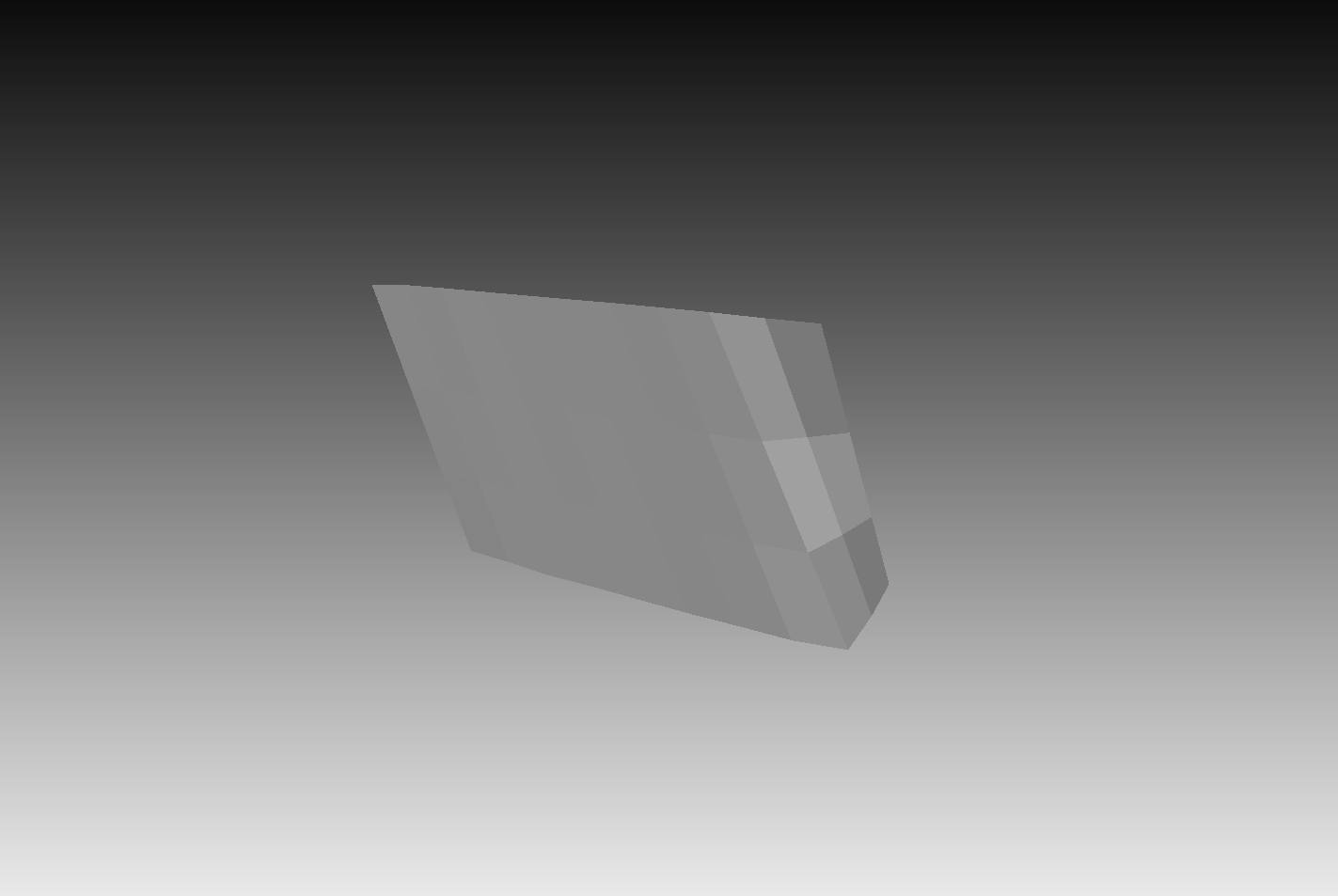}
    \includegraphics[width=0.325\linewidth]{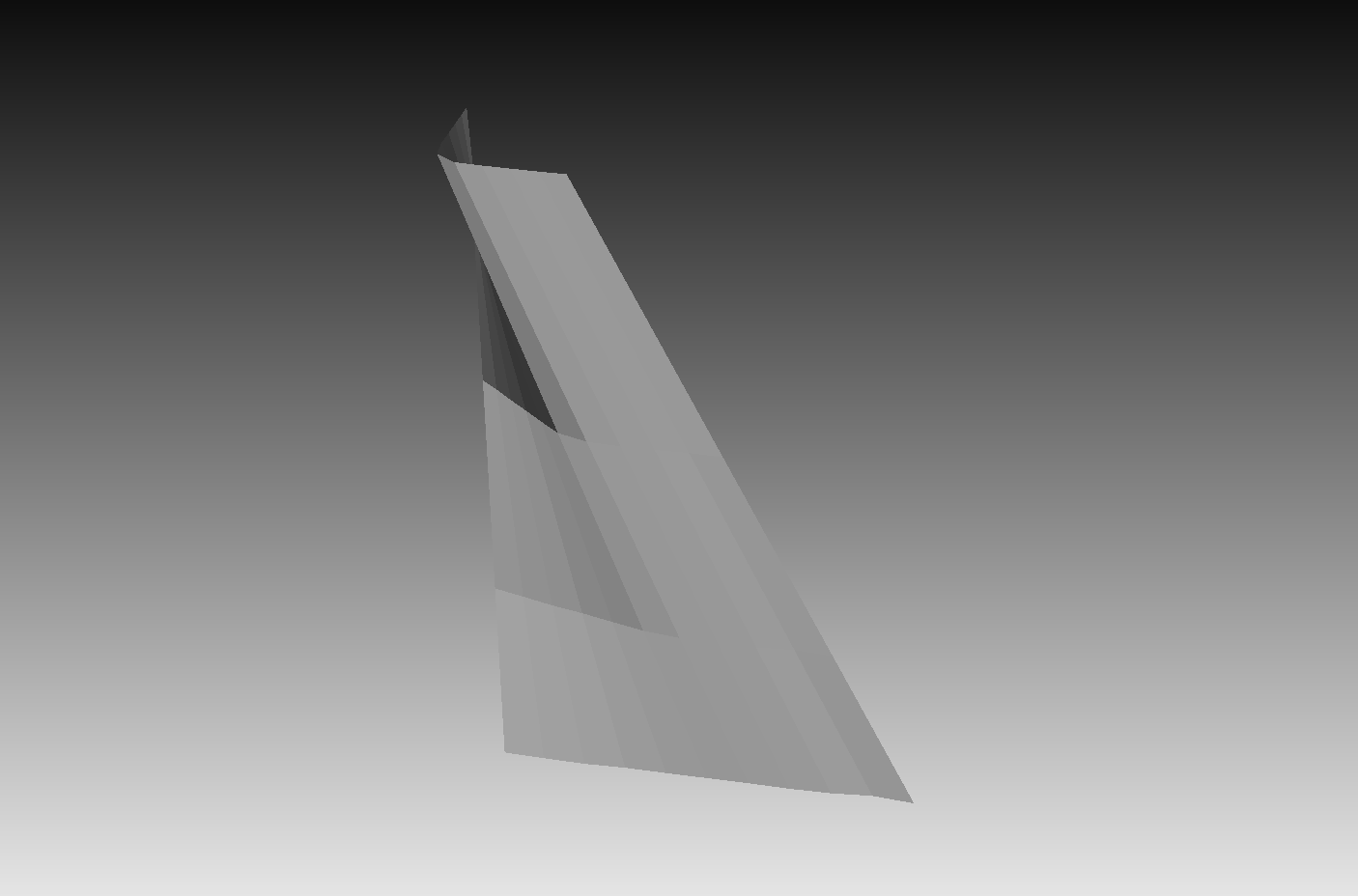}
    \includegraphics[width=0.325\linewidth]{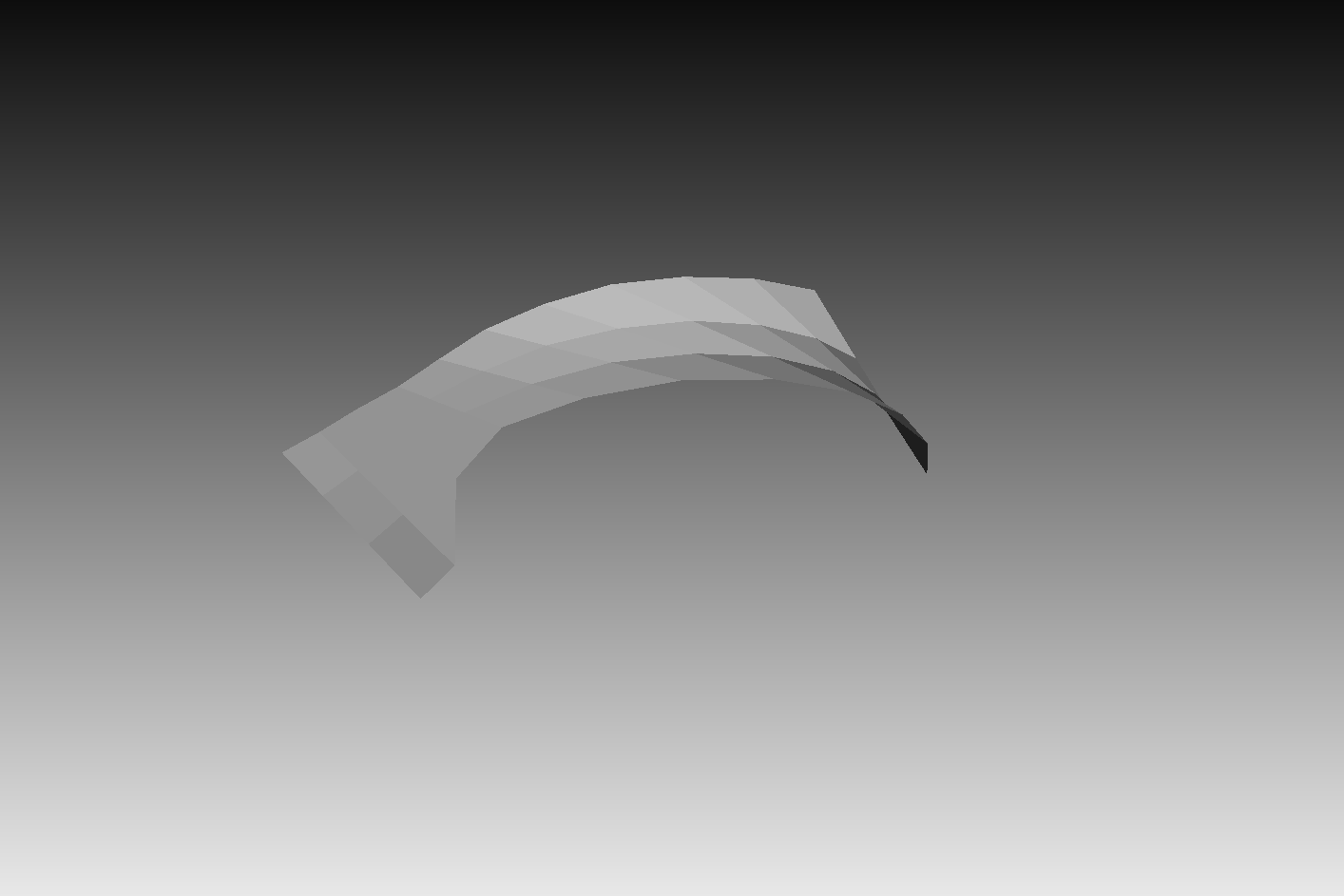}
    \includegraphics[width=0.325\linewidth]{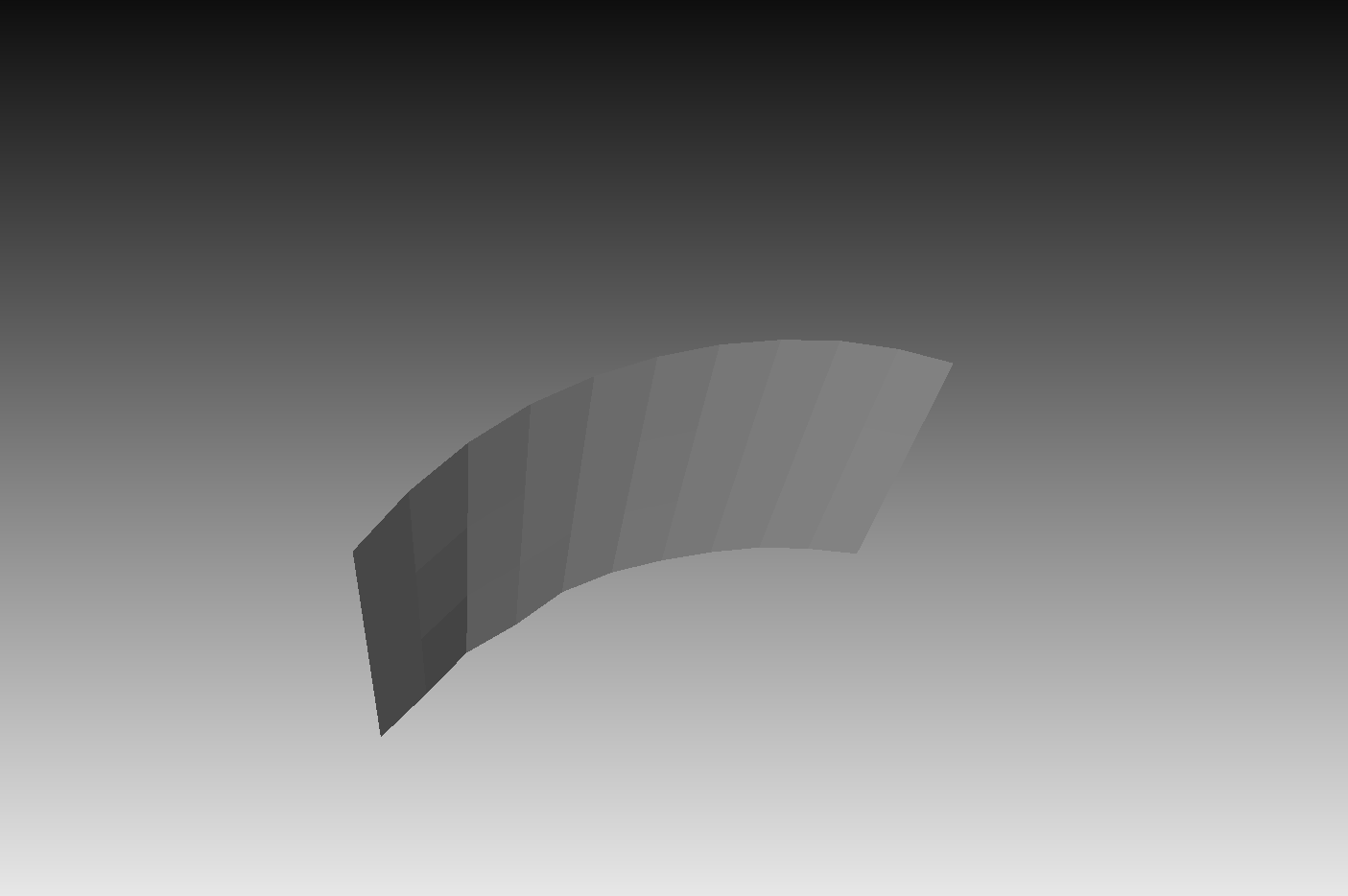}
    \includegraphics[width=0.325\linewidth]{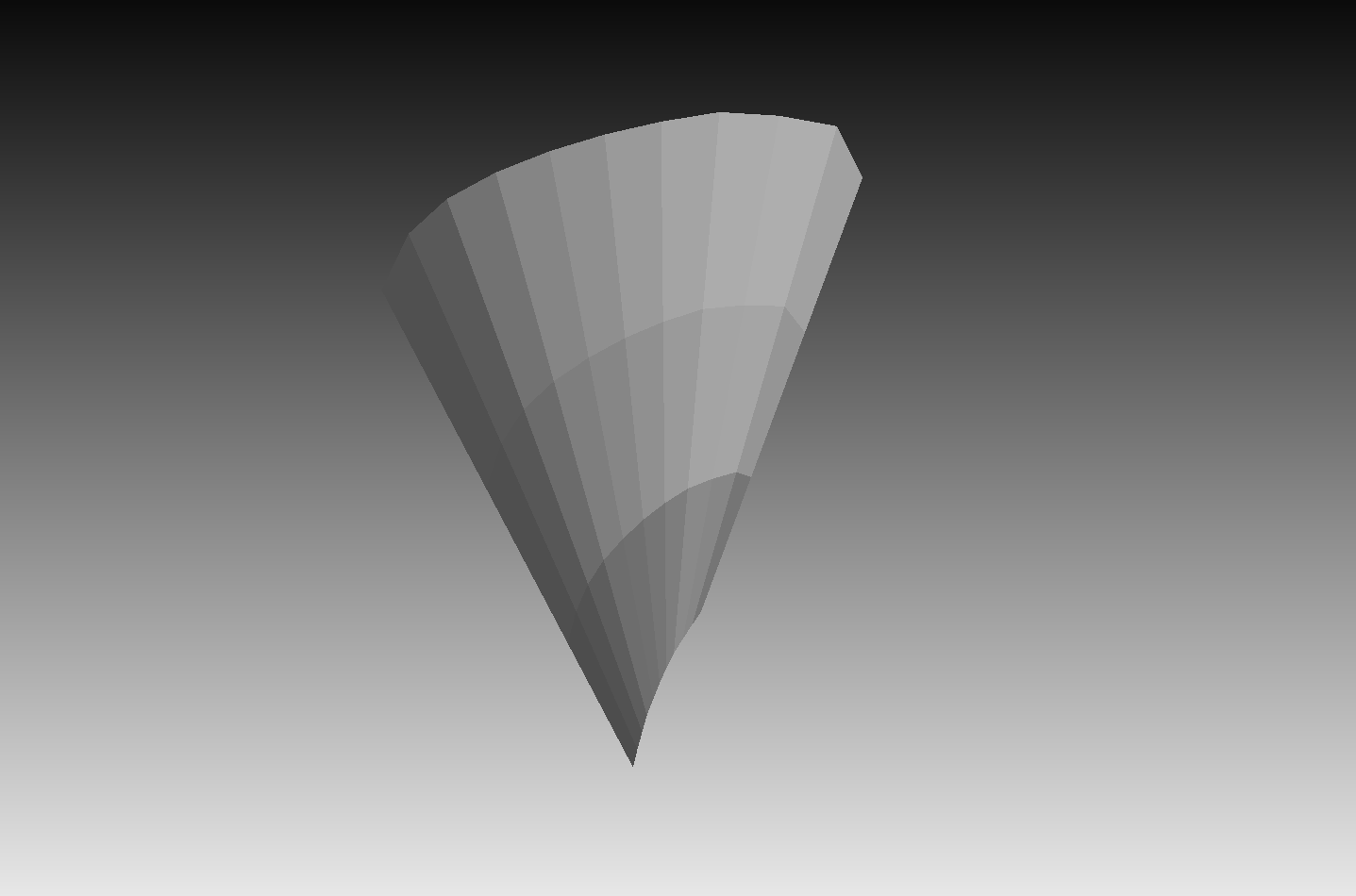}
  \end{center}
  \ReduceBeforeCaptionfigspace
  \caption{Some example loft surfaces of various geometries that our
    reconstruction algorithm generates.
  }
  \ReduceAfterCaptionfigspace
  \label{fig:lofting:samples}
\end{figure}

Note that an alternate method for forming pairs of 3D curve fragments is to use
the topology of 3D curve fragments as projected onto 2D views. The topology of
2D image curves is derived from the medial axis or Delaunay Triangulation to
determine the neighboring curve fragments for any given curve. The topology of
projected 3D curve fragments then induces a neighborhood relationship among 3D
curve fragments: two 3D curve fragments are neighbors in 3D if their
corresponding 2D image curves are neighbors in at least one view. This improves
the performance in two ways: {\em (i)} veridical pairing which exceed the
proximity threshold are restored to the pool of candidate pairs; {\em (ii)}
non-veridical curve pairs which are not neighbors are correctly discarded. This
is a significant factor in areas dense in 3D curves compared to the proximity
threshold, which generates numerous non-veridical curve pairs.

\noindent{\textbf{Hypothesis Viability Using Occlusion Consistency:}} The most
important cue in probing the viability of a 3D surface patch hypothesis is
whether it is consistent with respect to the occlusions it predicts (it is
assumed that surfaces are opaque). If an opaque 3D surface patch is veridical,
then all 3D curve structures that are occluded by it in a given projected image
must be invisible. For example, a surface hypothesis may occlude a portion of a
3D curve. Image evidence supporting the occluded portion is grounds for
invalidating the surface hypothesis, Figure~\ref{fig:occlusion:schematic}.

\begin{figure}
  \begin{center}
    \includegraphics[width=\linewidth]{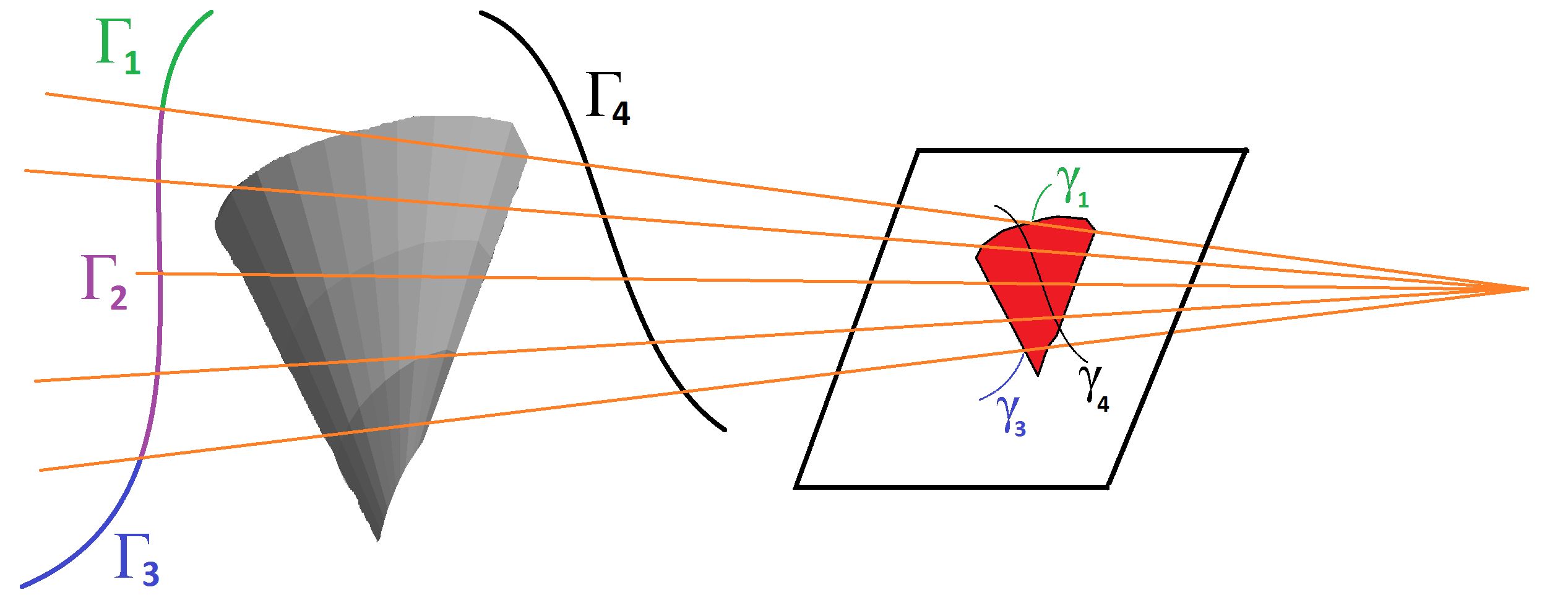}
  \end{center}
  \ReduceBeforeCaptionfigspace
  \caption{A 3D surface patch $S$ occludes all 3D curve fragments that lie
    behind it. Thus, the 3D curve fragments between $\Gama_1$ and $\Gama_4$ are
    partially obstructed so that only portions between $(\Gama_1, \Gama_2)$ and
    $(\Gama_3, \Gama_4)$ are visible as $(\gama_1, \gama_2)$ and $(\gama_3,
    \gama_4)$ in the image. The projections of $(\Gama_2, \Gama_3)$ should have
    no edge evidence in the image. On the other hand, the 3D curve fragments
    $(\Gama_5, \Gama_6)$ is fully unoccluded and edge evidence for it is
    expected. The presence of edge evidence in the portion $(\gama_2, \gama_3)$
    is grounds for invalidating the 3D surface hypothesis $S$.
  }
  \ReduceAfterCaptionfigspace
  \label{fig:occlusion:schematic}
\end{figure}

The technical approach to testing occlusion is based on ray tracing
\cite{Glassner:raytracing:book}: A ray is connected from the camera center to
each point on a 3D curve fragment belonging to the 3D curve drawing and the
visibility of the point is tested against each surface hypothesis. Specifically,
let $\{\Pi_1, \dots, \Pi_N\}$ denote the set of hypothesized surface patches.
Let the 3D curve drawing have curve fragments $\{\Gama_1, \dots, \Gama_K\}$,
each having image curve projections onto view $l$, $\gama_k^l (s)$, where $s$
represents length parameter $s \in [0,L_k^l]$, where $L_k^l$ is the total length
of the projected curve. Let the portion of the 3D curve that is occluded by the
surface patch $\Pi_n$ be denoted by the interval $(a_{k,n}^l, b_{k,n}^l)$. Then,
the evidence against surface hypothesis $\Pi_n$ provided by curve $\Gama_k$ from
view $l$, $E_{k,n}^l$, is the edge support for the invisible portion. This
evidence is the sum of total edge support at sample point $s$, $\phi(\gama_k^l
(s))$, which is simply the number of image edges that have matching locations
and orientations to the curve $\gama_k^l (s)$ at sample point $s$:

\begin{equation}\label{eq:lofting:support}
E_{n,k}^l = \int_{a_{k,n}^l}^{b_{k,n}^l} \phi(\gama_k^l (s))ds
\end{equation}

This evidence is then subjugated to a threshold of significance $\tau_E$; if
significant, the evidence invalidates the hypothesis. On the other hand, if the
evidence against the hypothesis for all the curves that should be occluded is
indeed insignificant, \ie, $E_{n,k}^l < \tau_E, \forall k$, the lack of evidence
in fact provides support for the surface hypothesis. This is to be distinguished
from surface hypotheses that are not occluding any curves. The situation where
$\Pi_n$ occludes $\Gama_k$ and image evidence shows occlusion lends more
evidence to $\Pi_n$ than the situation where $\Pi_n$ does not occlude any
curves.

We now assume that all surface patches occlude at least one curve in at least
one view; note that for polyhedral shapes, frontal patches occlude the contours
of patches on the back, so this is not a stringent assumption. In fact, probing
this assumption on both Amsterdam House Dataset and Barcelona Pavilion Dataset
(which are described in Section~\ref{sec:lofting-results}) shows that this is
the case for more than 90\% of the surface hypotheses generated. This assumption
implies that each surface hypothesis needs to be confirmed at least once against
an occlusion hypothesis, \ie, $\forall n, \exists l, \exists k, \text{such that
} E_{k,n}^l < \tau_E$.

The above process probes the implication of surface patch in relation to the 3D
curve drawing. When introducing a multitude of surface patches, however, the
issue of occlusion between two surface hypotheses arises. It is possible that
one surface hypothesis is fully occluded by all other surfaces. Such a surface
is then not visible in any view and is discarded.

\begin{figure*}
  \begin{center}
    \includegraphics[width=\linewidth]{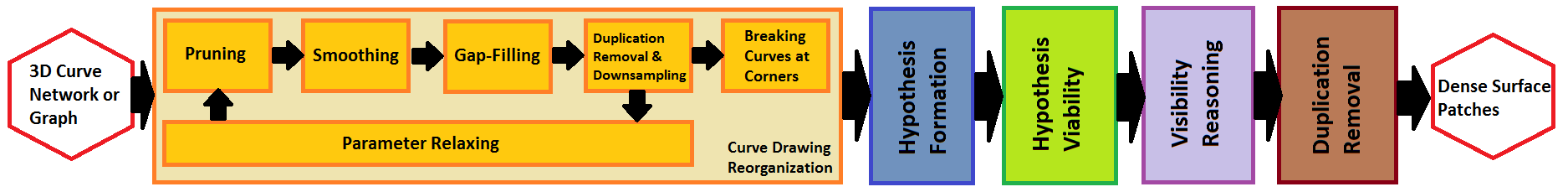}
  \end{center}
  \caption{A visual illustration of our dense surface reconstruction pipeline.
  }
  \label{fig:lofting:pipeline}
\end{figure*}

\noindent{\textbf{Redundant Hypotheses:}} Since surface hypotheses are generated
by pairs of 3D curve fragments, if a ground truth surface consists of multiple
curve fragments, say a rectangular patch consisting of four curve fragments,
then the same surface will likely be represented by a number of curve fragment
pairs, six possible pairs in the case of a rectangular patch. 

These redundant representations are detected in a post-processing stage and
consolidated. When a large portion of a surface hypothesis (80\% in our system)
is subsumed by another surface, \ie, 80\% of the points on it are closer than a
proximity threshold to another surface, then this surface is discarded as a
redundant hypothesis. A more principled approach is to merge two overlapping
surfaces by forming curve triplet hypotheses: When two curve pairs have a curve
fragment in common and their surface hypotheses overlap, as described above, the
lofting approach is applied to the curve triplet and the resulting surface
replaces the pair of surface hypotheses. And, of course, a curve triplet and a
curve pair with a common curve fragment and overlapping surfaces result in curve
quadruplet hypotheses, and so on as needed. This growth of surface hypotheses
yields more accurate and less redundant surface patches, but results from this
process are not ready for inclusion in this publication.

\begin{figure}
  \begin{center}
    \includegraphics[width=0.485\linewidth]{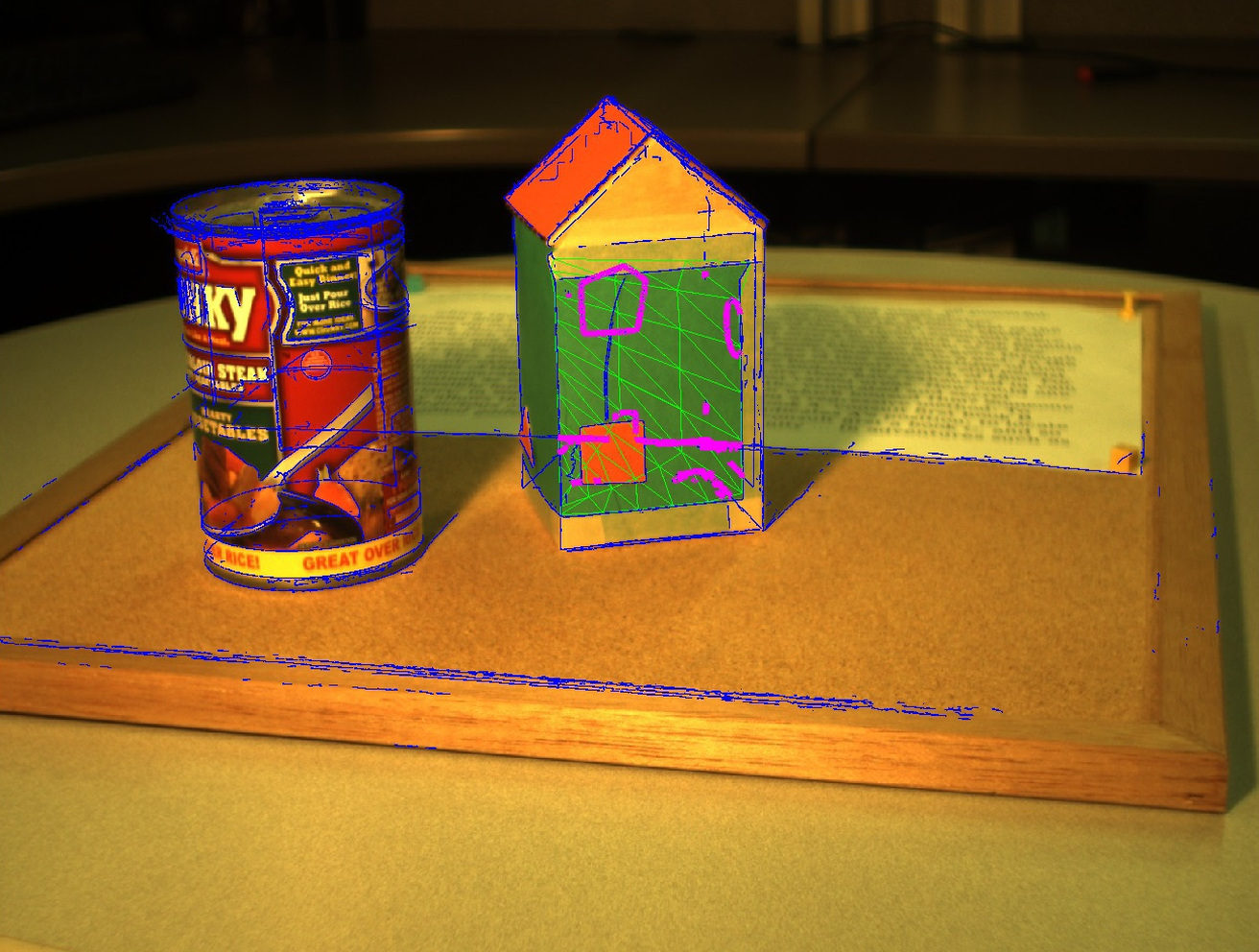}\hspace{1.3mm}\includegraphics[width=0.485\linewidth]{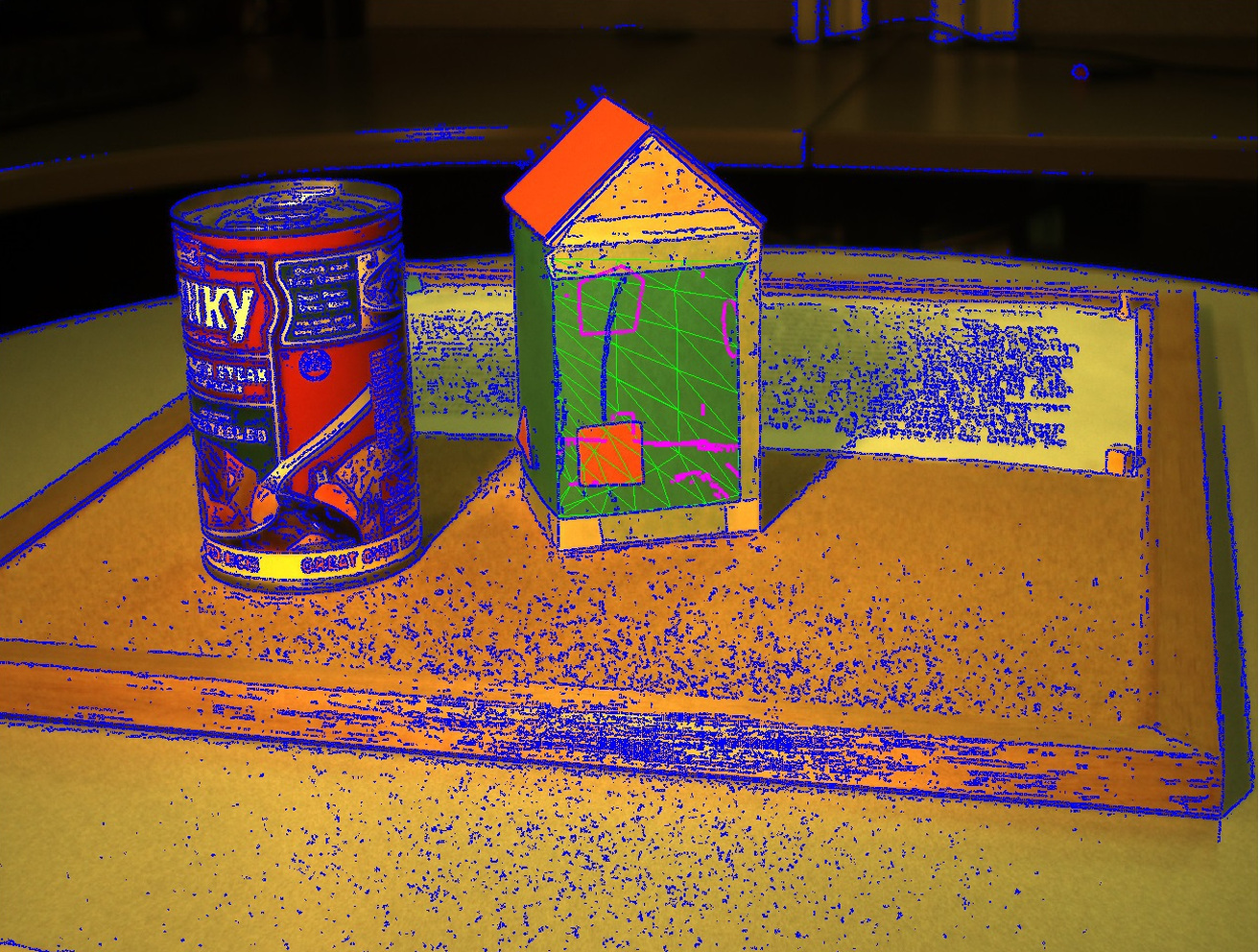}\\[1mm]
\includegraphics[width=0.485\linewidth]{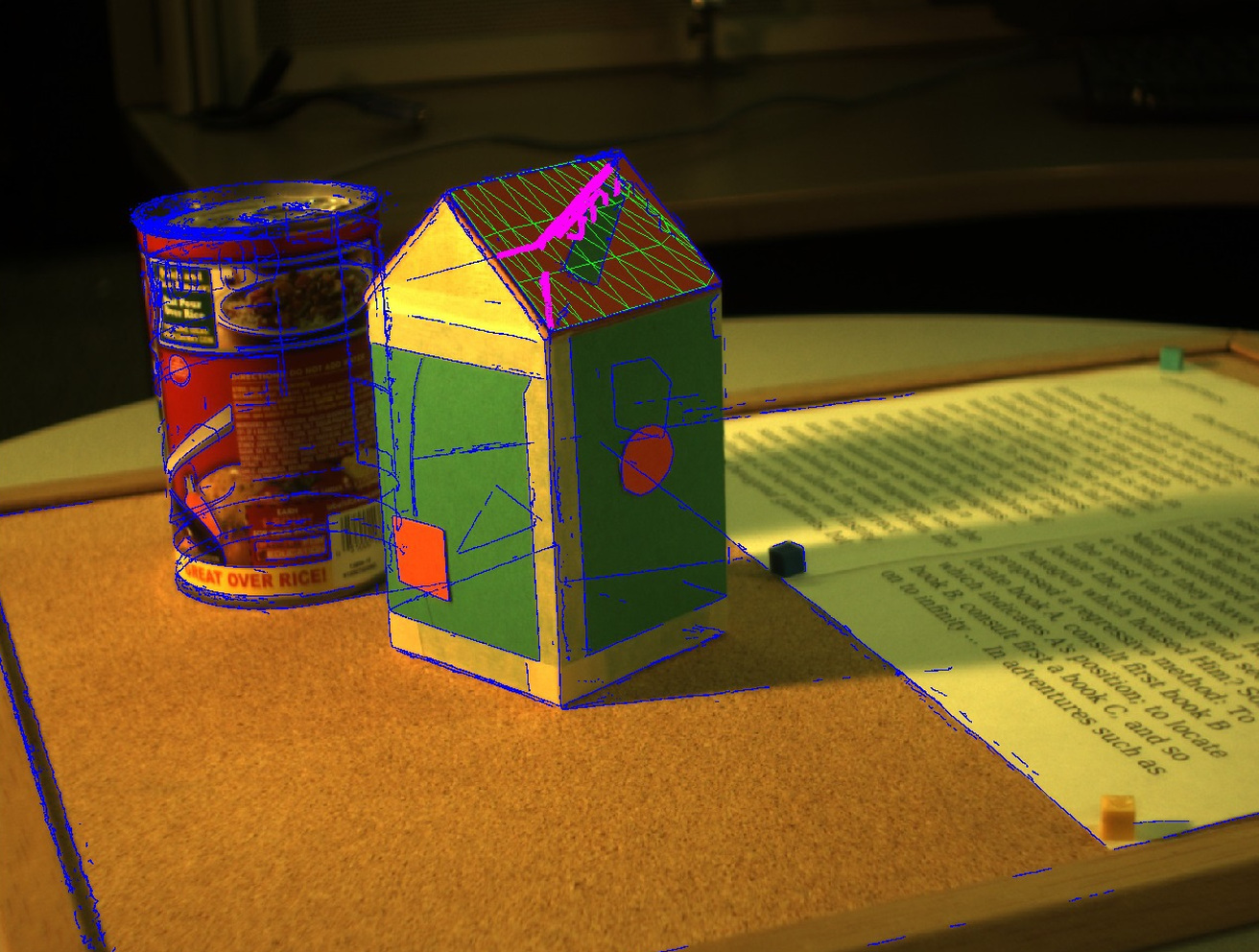}\hspace{1.3mm}\includegraphics[width=0.485\linewidth]{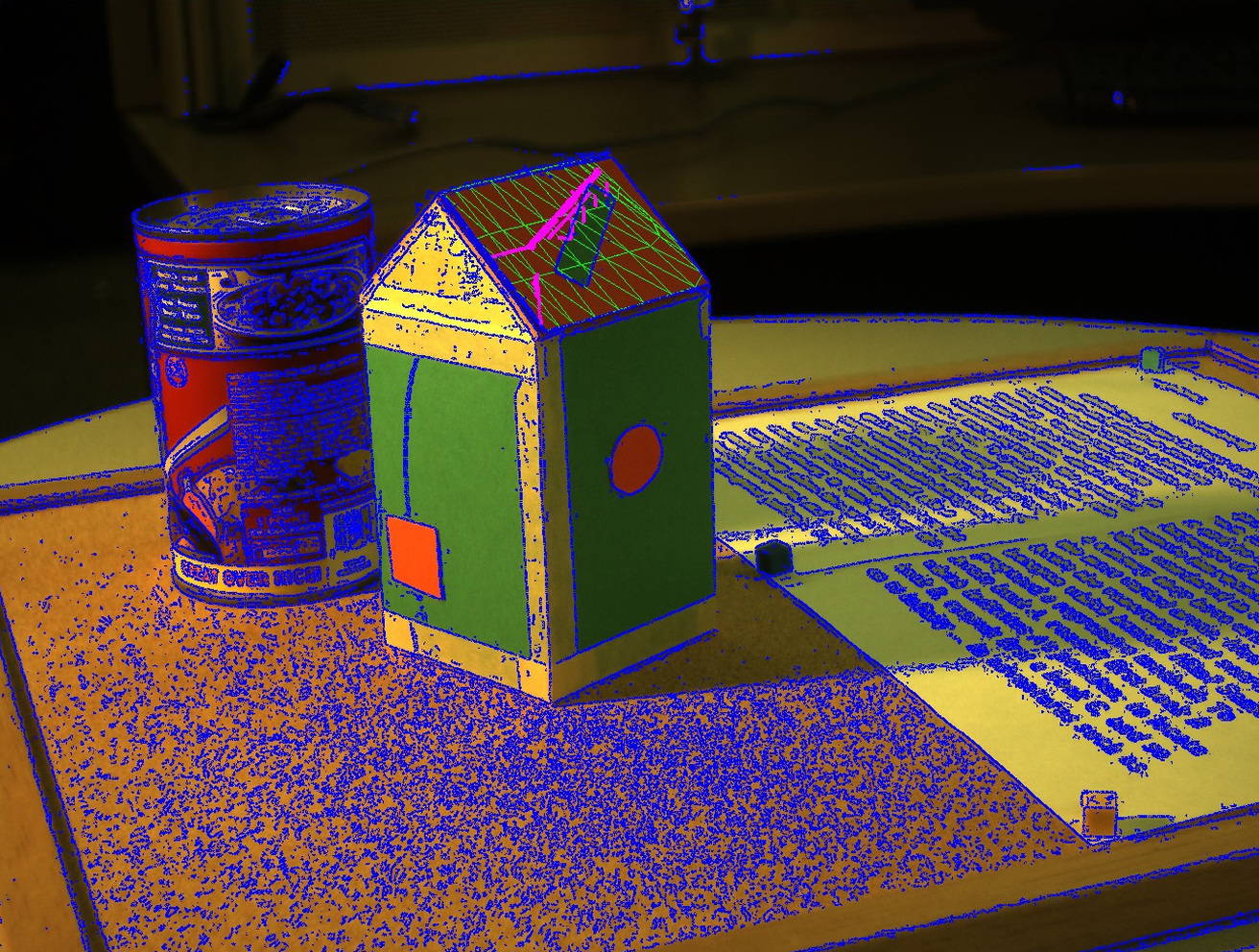}
  \end{center}
  \ReduceBeforeCaptionfigspace
  \caption{Examples of surface hypotheses being confirmed by the confirmation
    views shown here. 
    Left column: Projected surface hypothesis is shown in green, projected curve
    drawing is shown in blue and occluded segments are shown in purple. Right
    column: Same surface and occluded segments are shown with image edges in
    blue. Notice the lack of any edge presence whatsoever around most of the
    purple segments, which is clear indication of occlusion consistency between
    the images and the hypothesis surface.
  }
  \ReduceAfterCaptionfigspace
  \label{fig:confirm}
\end{figure}

Figure~\ref{fig:lofting:pipeline} is a visual illustration of our entire surface
reconstruction approach. Figure~\ref{fig:confirm} demonstrates that our
algorithm is very good at correlating image edges with 3D curve structures,
accurately reasoning about occlusion and confirming an overwhelming majority of
correct surfaces, as well rejecting almost all of the incorrect hypotheses,
Figure~\ref{fig:false}. It should be noted that many surface hypotheses do not
contain any portion of the curve drawing behind them from {\em any} given view.
These hypotheses cannot be confirmed or denied, and, depending on the robustness
of the hypothesis generation algorithm, they can be included in or discarded
from the output as needed. In addition, many existing multiview stereo methods
can be plugged into our system at the level of curve pairing and used as
alternative ways to provide initial seeds for our surface hypotheses. As
mentioned earlier, our lofting algorithm scales well to a large number of input
3D curves, which are provided either simultaneously or sequentially.

\begin{figure}
  \begin{center}
    \includegraphics[width=0.485\linewidth]{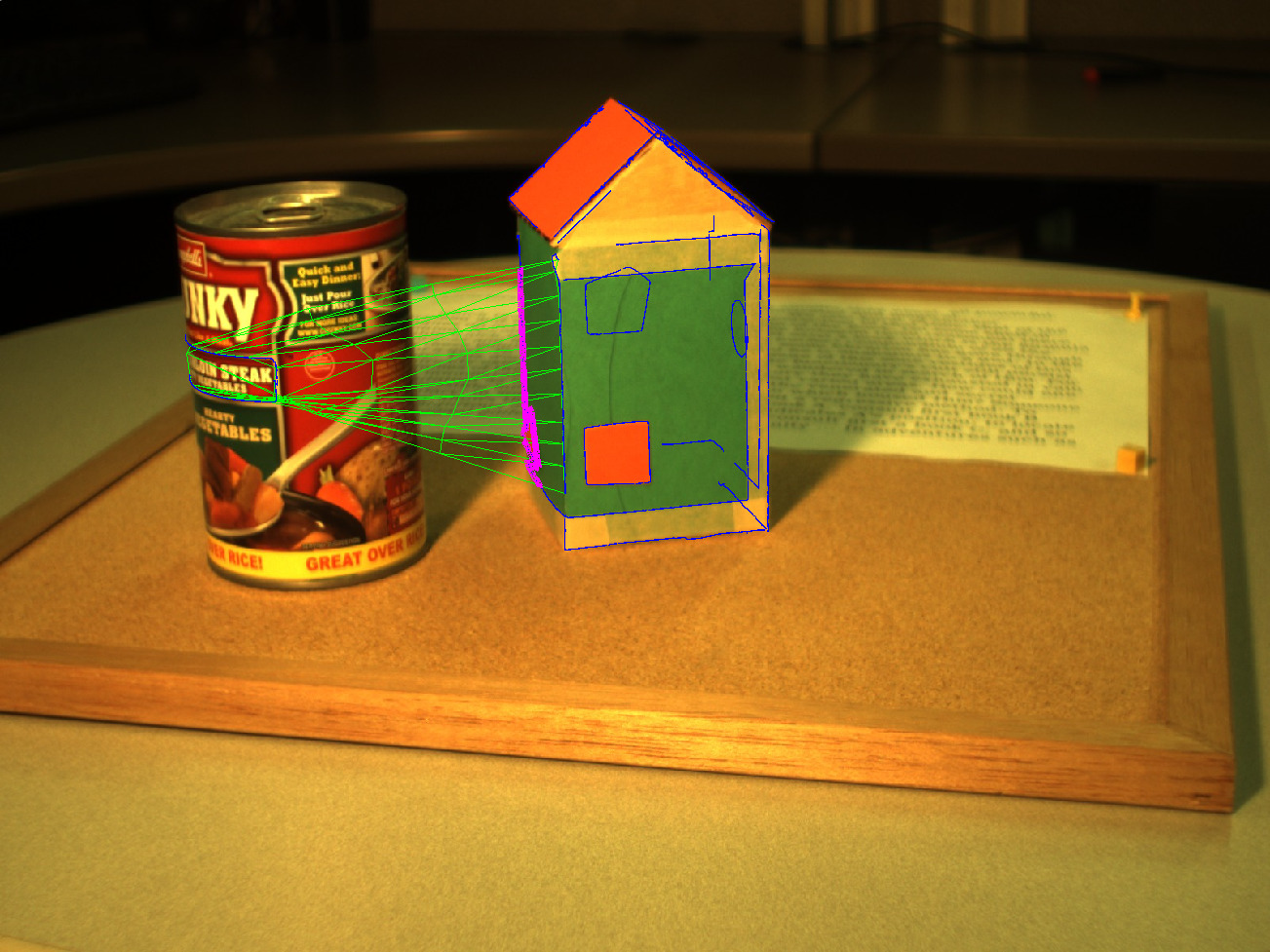}\hspace{1.3mm}\includegraphics[width=0.485\linewidth]{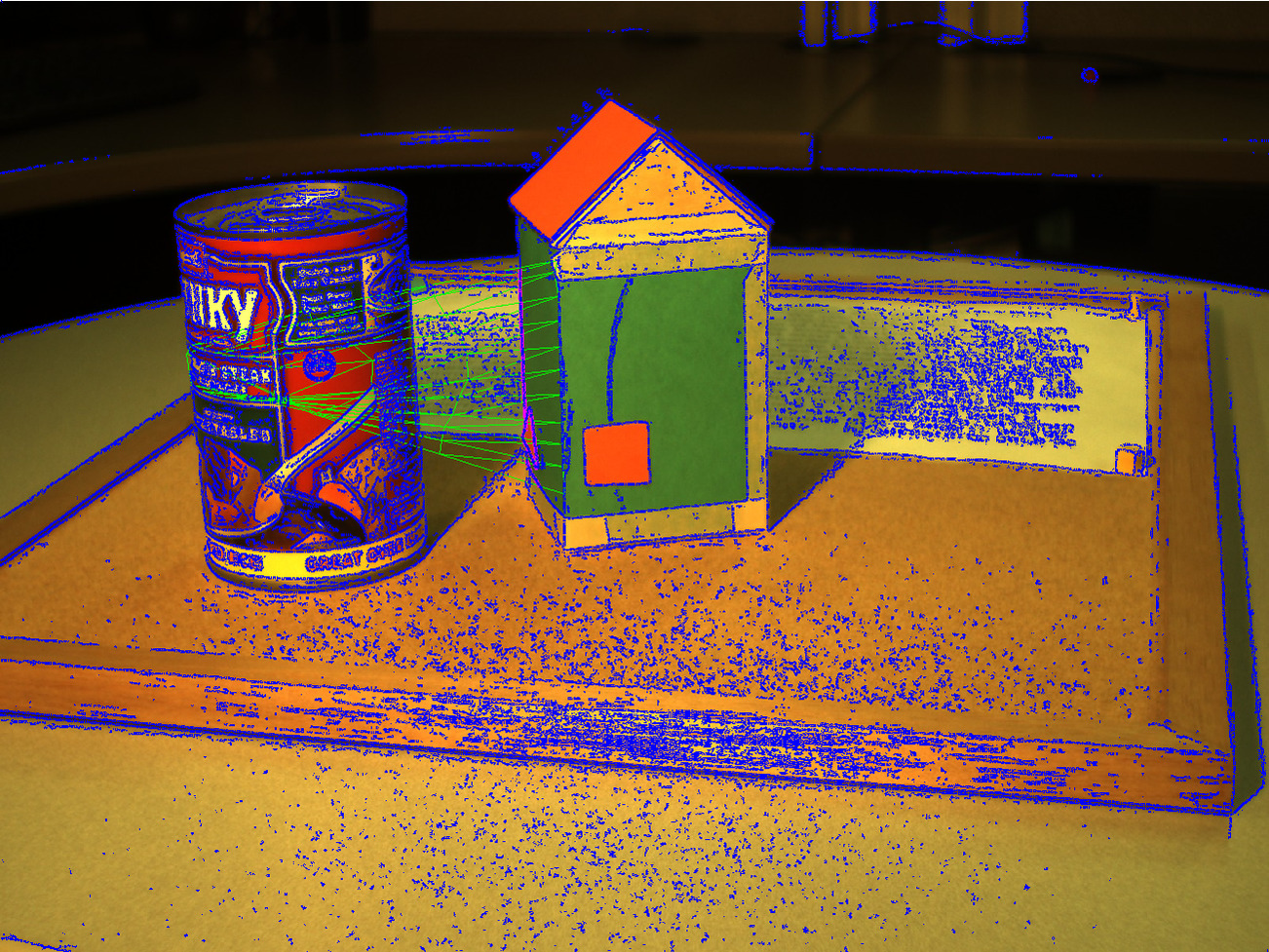}\\[1mm]
\includegraphics[width=0.485\linewidth]{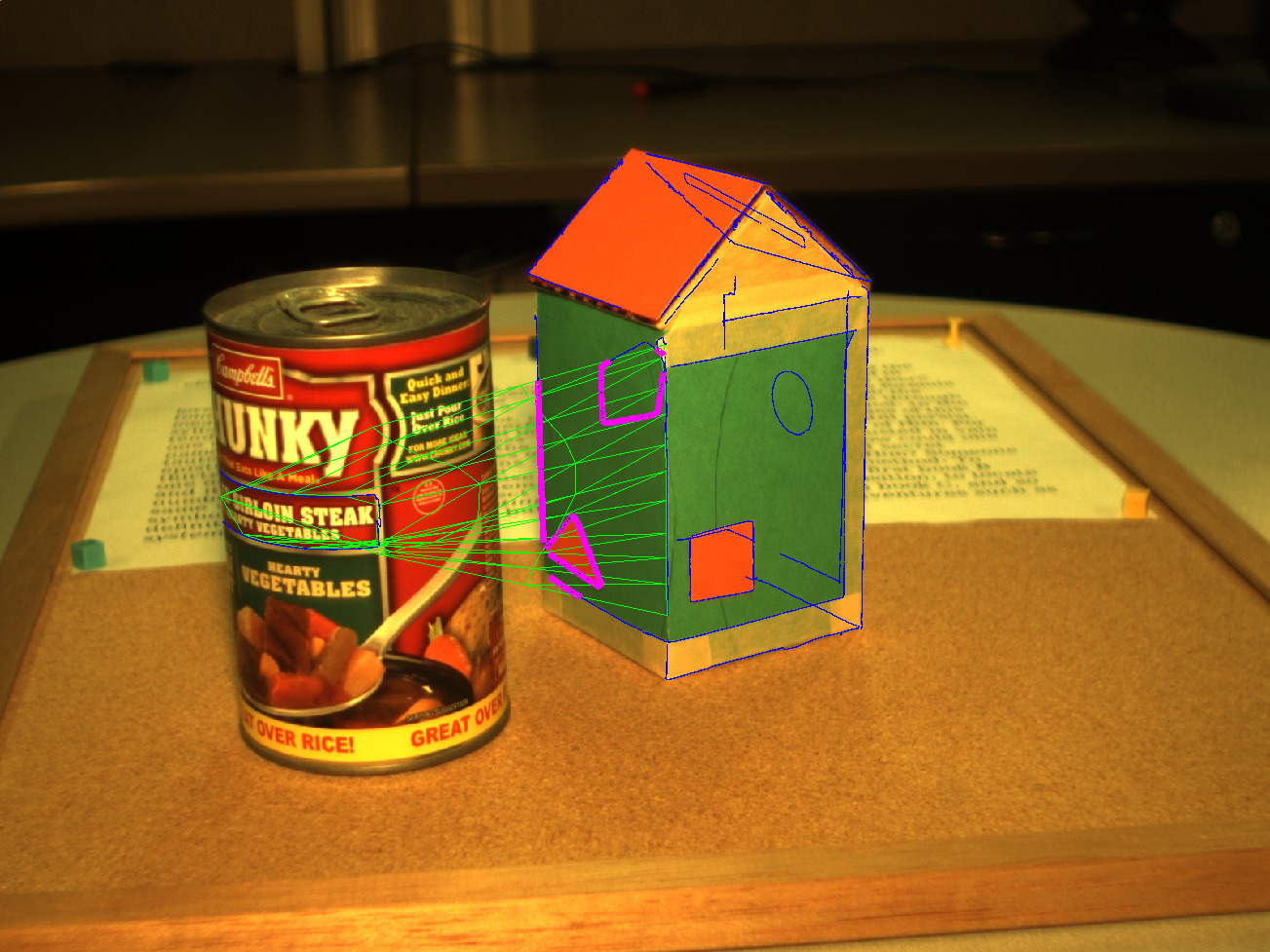}\hspace{1.3mm}\includegraphics[width=0.485\linewidth]{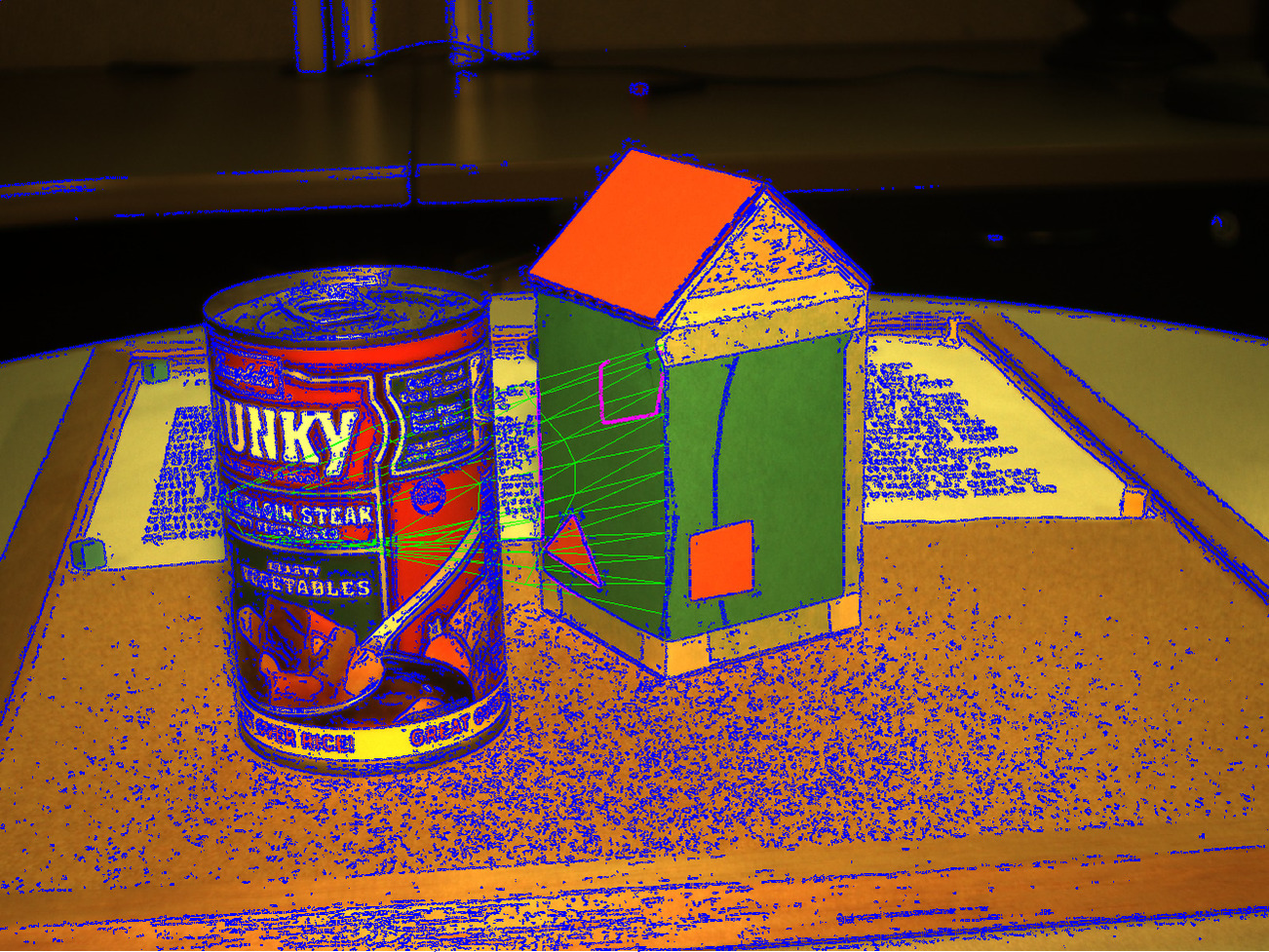}
  \end{center}
  \ReduceBeforeCaptionfigspace
  \caption{An example outlier surface hypothesis ruled out by detected edge
    structures. Left column: projected surface hypothesis is shown in green,
    projected curve drawing is shown in blue and occluded segments are shown in
    purple. Right column: Same surface and occluded segments are shown with
    image edges in blue. Notice how most of the purple segments are barely
    visible from all the edges that match in both location and orientation.
  }
  \ReduceAfterCaptionfigspace
  \label{fig:false}
\end{figure}

\section{Reorganization of Input Curve Graph Using Differential Geometric Cues}
\label{sec:reorg}

Four important technical issues arise in the application of lofting to
reconstruct surface patches from 3D drawings. 

\noindent{\textbf{\underline{Problem 1:} Lofting sensitivity to overgrouping:}}
Lofting is highly sensitive to overgrouping of edges into curves. 
If some parts of a curve belong to a veridical surface patch but another part
does not, then the lofting results experience significant and irreversible
geometric errors, \eg, as in Figure~\ref{fig:lofting:sensitivities}a where two
curve fragments $\mathcal{C}_1$ and $\mathcal{C}_2$ belong to a side of the
house and correctly hypothesize a surface patch through lofting. However, if
$\mathcal{C}_2$ is grouped with an adjacent curve fragment $\mathcal{C}_3$
belonging to an adjacent face of the house that $\mathcal{C}_2$ belongs to (let
$\mathcal{C}_4$ denote $\mathcal{C}_2 \cup \mathcal{C}_3$), then the lofting
results based on $(\mathcal{C}_1$, $\mathcal{C}_4)$ do not produce a meaningful
surface patch. The core of this problem is that the curve $C_2$ is shared by two
surface hypotheses, but if grouped with $C_3$, it can no longer represent the
frontal surface hypothesis created by $C_1$ and $C_2$. This transition in the
ability to represent multiple surface hypotheses happens at junctions. Thus,
breaking all curves at corners, \ie high-curvature points, should remedy this
problem, Figure~\ref{fig:reorg:results}.  Unfortunately, it is difficult to
output curvature for noisy curves, thus requiring a smoothing algorithm before
the curve can be broken at high-curvature points. This smoothing algorithm is
described below in the context of curve noise.

\noindent{\textbf{\underline{Problem 2:} Lofting sensitivity to curve noise: }}
Curve fragments of the 3D drawing can have excessive noise, depicting
loop-like structures and local perturbations,
Figure~\ref{fig:lofting:sensitivities}b. These degeneracies in the local form of
a curve fragment often result in failures in the lofting algorithm to produce a
surface hypothesis, or result in surfaces featuring geometric degeneracies.
There are a number of smoothing methods, and we use a relatively recent robust
algorithm that is based on B-splines~\cite{Garcia:CSDA2010, Garcia:EIF2011},
balancing data fidelity term with a smoothness term. The ratio of those two
terms determine the degree of smoothing. The advantage of this method is that
the polyline representation of the curve can be maintained after smoothing.

\noindent{\textbf{\underline{Problem 3:} Lofting sensitivity to
overfragmentation and gaps: }} Lack of edges or undergrouping in the edge
grouping stage can lead to gaps and  overfragmentation. In both cases, a long
veridical curve is represented as multiple smaller curve fragments,
Figure~\ref{fig:lofting:sensitivities}c. As a result, what would have been a
single surface patch now needs to be covered by a suboptimal set of smaller,
overlapping surface hypotheses. In addition, the increased number of curve
fragments increases the number of curve pairs to be considered, and lead to a
combinatorial increase in computational cost.  Curve fragments that are
coincidental at a point can be grouped if they show good continuity of tangents
at endpoints. Similarly, gaps between two curve fragments $\Gama_1(s)$ and
$\Gama_2(s)$ can be bridged between endpoint $\Gama_1(s_1)$ and $\Gama_2(s_2)$
if: {\em (i)} These endpoints are sufficiently close, \ie, $|\Gama_1(s_1) -
\Gama_2(s_2)| < \tau_{dist}$, where $\tau_{dist}$ is a gap proximity threshold,
and {\em (ii)} $CC((\Gama_1(s_1),T_1(s_1)),(\Gama_2(s_2),T_2(s_2))) <
\tau_{cocirc}$ where $CC$ is the co-circularity measure, characterizing good
continuation from one point-tangent pair $(P_1,T_1)$ to another pair $(P_2,T_2)$
\cite{Parent:Zucker:PAMI1989}.

\begin{figure}
  \begin{center}
    \includegraphics[width=\linewidth]{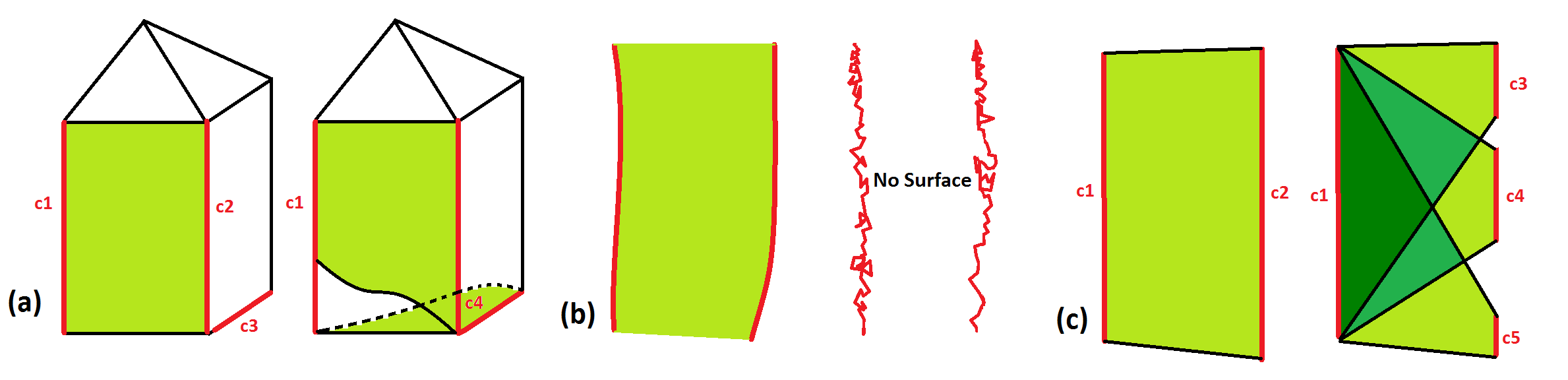}
  \end{center}
  \ReduceBeforeCaptionfigspace
  \caption{{\em (a)} Overgrouping of two curve fragments $\mathcal{C}_2$ and
  $\mathcal{C}_3$ into $\mathcal{C}_4$ can lead to nonsensical lofting results
  in the pair $(\mathcal{C}_1,\mathcal{C}_4)$ in contrast to the
  close-to-veridical results of lofting $(\mathcal{C}_1,\mathcal{C}_2)$; {\em
  (b)} lofting is sensitive to loop-like noise or excessive perturbations; {\em
  (c)} lofting with overfragmented curves produces suboptimal lofting results
  and redundant surface proposals, leading to a combinatorial increase in the
  number of lofting applications and postprocessing.  
  }
  \ReduceAfterCaptionfigspace
  \label{fig:lofting:sensitivities}
\end{figure}

\noindent{\textbf{\underline{Problem 4:} Duplications due to curve fragment
overlaps: }} There is some duplication in 3D curve fragments in that two curves
can overlap along portions, thus creating duplicate surface representations.
While this duplication may not be an issue for some applications, better results
can be obtained if the duplication is removed: When two curves overlap, the
longer curve is unaltered and the overlapping segment is removed from the
shorter curve. The curves are also downsampled since the initial curve drawing
is dense in sample points.

The resolution of the above four problems significantly improves the performance
of our algorithm. Note that these steps are applied in sequence: Pruning small
curves, smoothing curve fragments, gap filling and grouping overfragmented
segments, eliminating duplications and downsampling. In addition, it is
judicious to iteratively apply these steps in sequence, starting with small
parameters and increasing the parameters in steps (typically 3-4). This is
crucial because all of these steps run the risk of distorting the 3D data in
significant ways if pursued too aggressively in a single iteration, \eg, corners
can be oversmoothed, wrong gaps can be filled, meaningful but relatively short
curve fragments can get pruned without getting a chance to be merged into a
larger curve fragment etc.

It should be noted that aforementioned problems do not arise in the plethora of
interactive surface lofting approaches, as a human agent is available to break
or group 3D structures to obtain geometrically accurate 3D surfaces,
\cite{Nealen:etal:SIGGRAPH2007}. Some of the lofting approaches try to get
around this problem by constraining the input curves to be closed curves,
\cite{Zhuang:etal:ACMTOG2013, Schaefer:etal:SGP2004}, but a fully automated,
bottom-up lofting system like ours has to be able to handle such grouping
inconsistencies algorithmically. 

\begin{figure} 
  \begin{center}
    \includegraphics[width=0.49\linewidth]{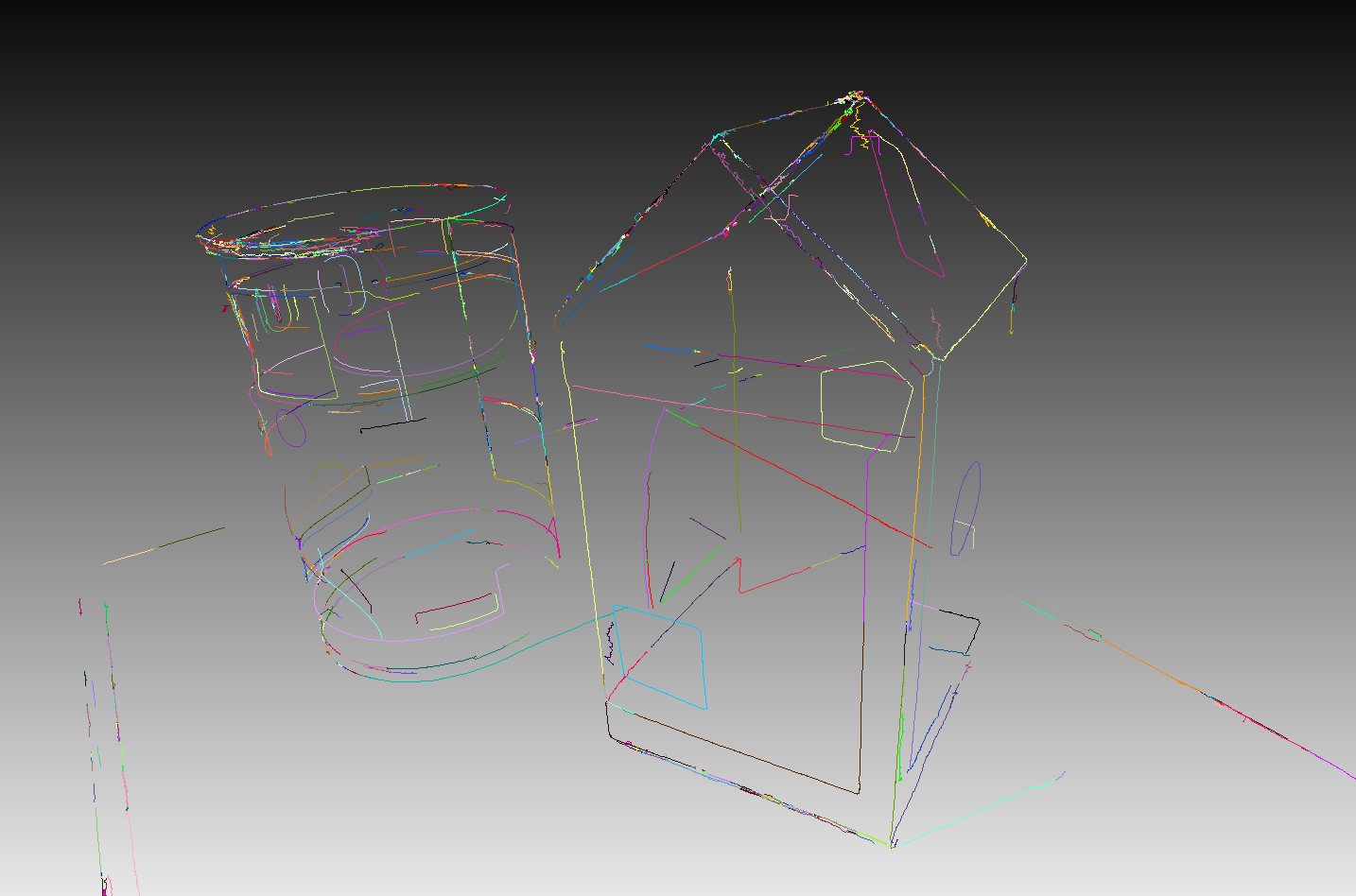}
    \includegraphics[width=0.49\linewidth]{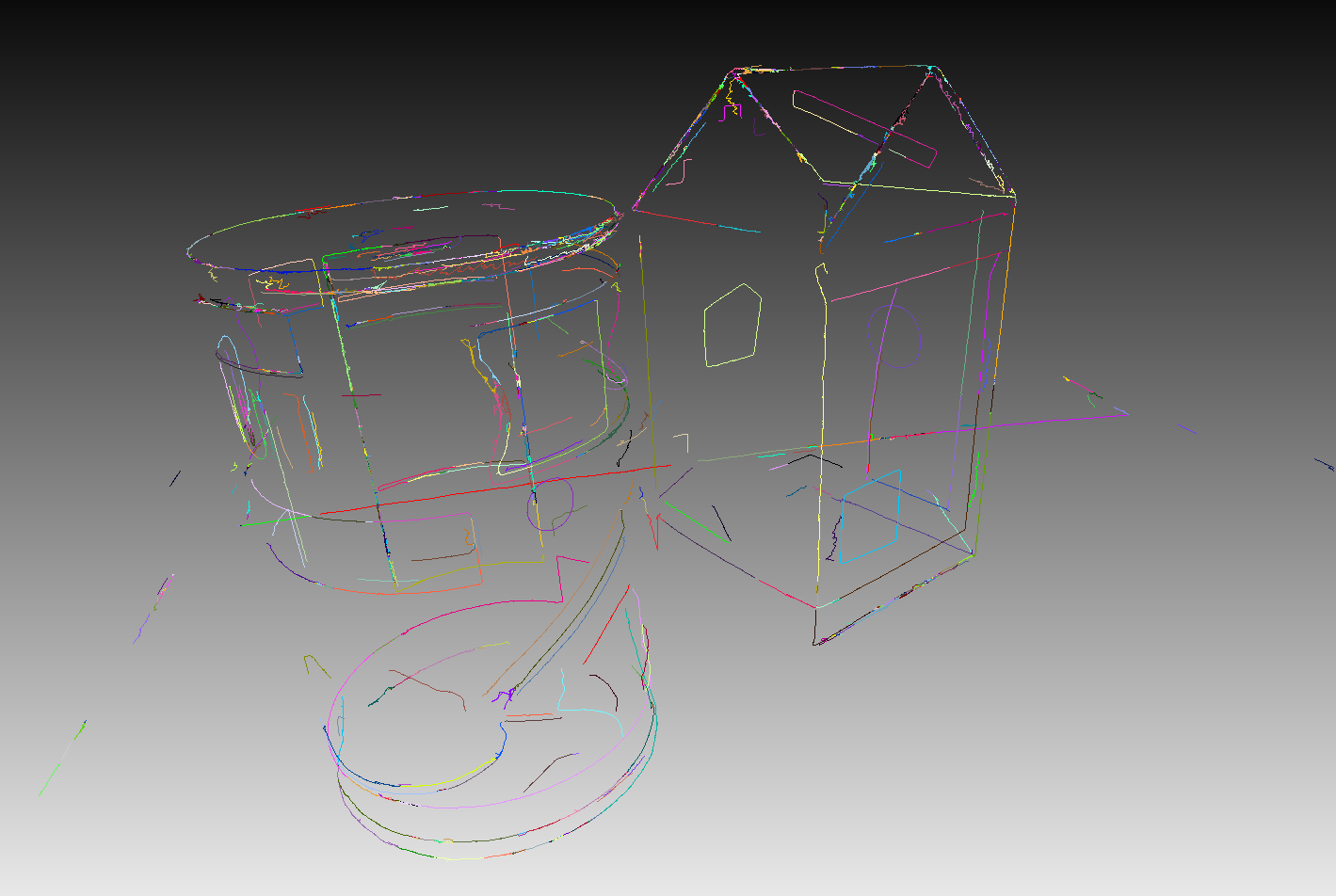}
    \includegraphics[width=0.49\linewidth]{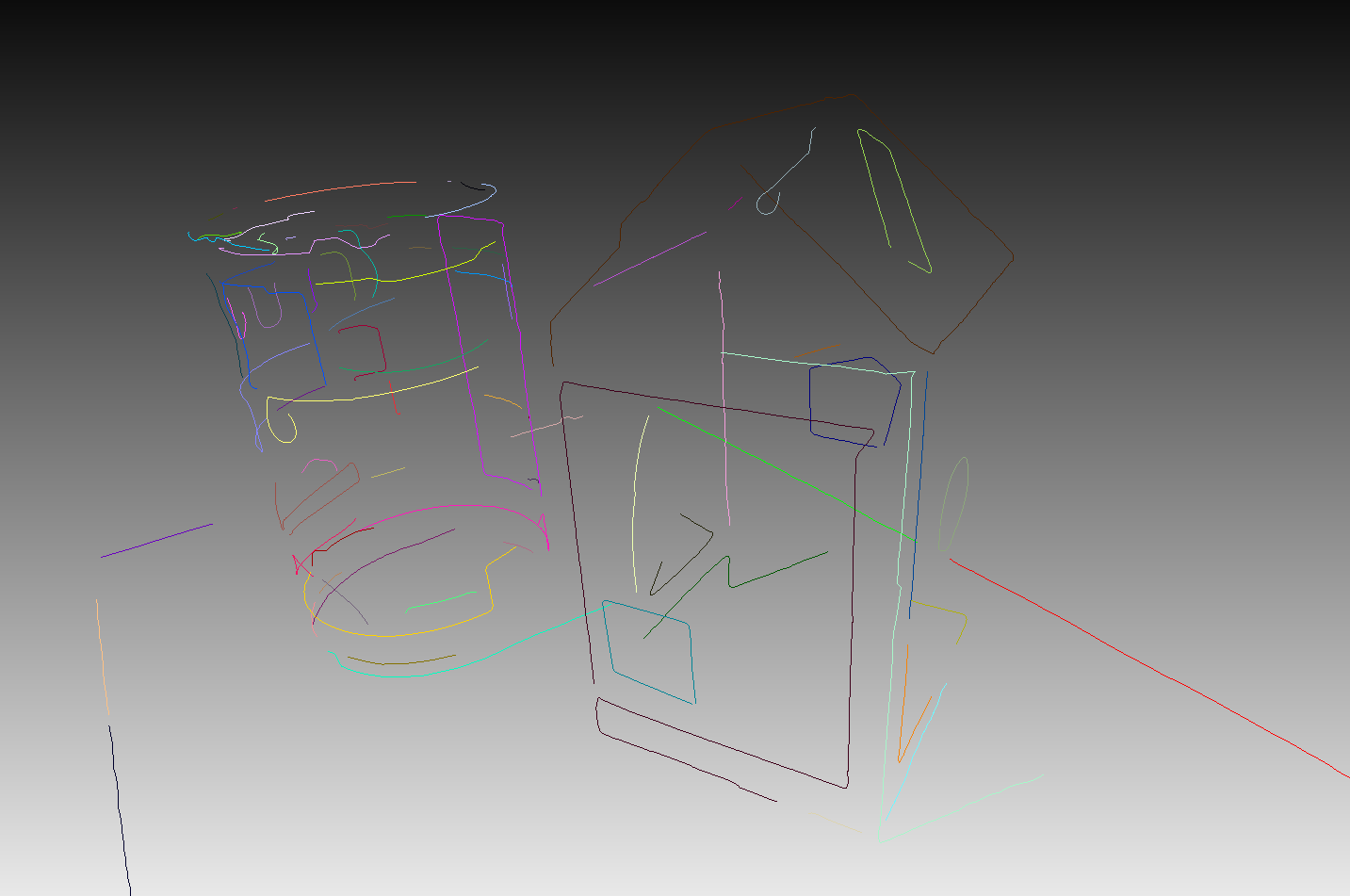}
    \includegraphics[width=0.49\linewidth]{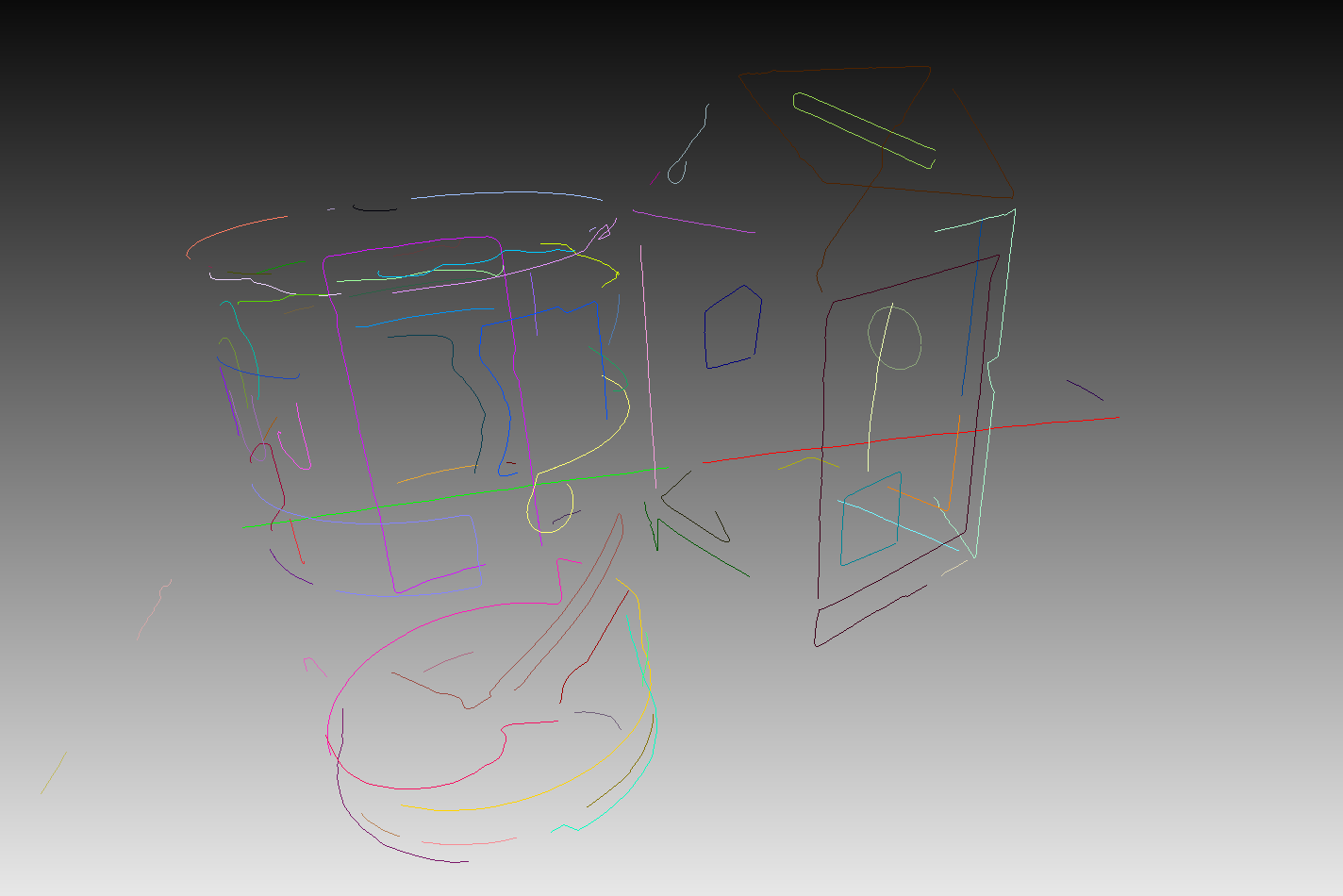}
    \includegraphics[width=0.49\linewidth]{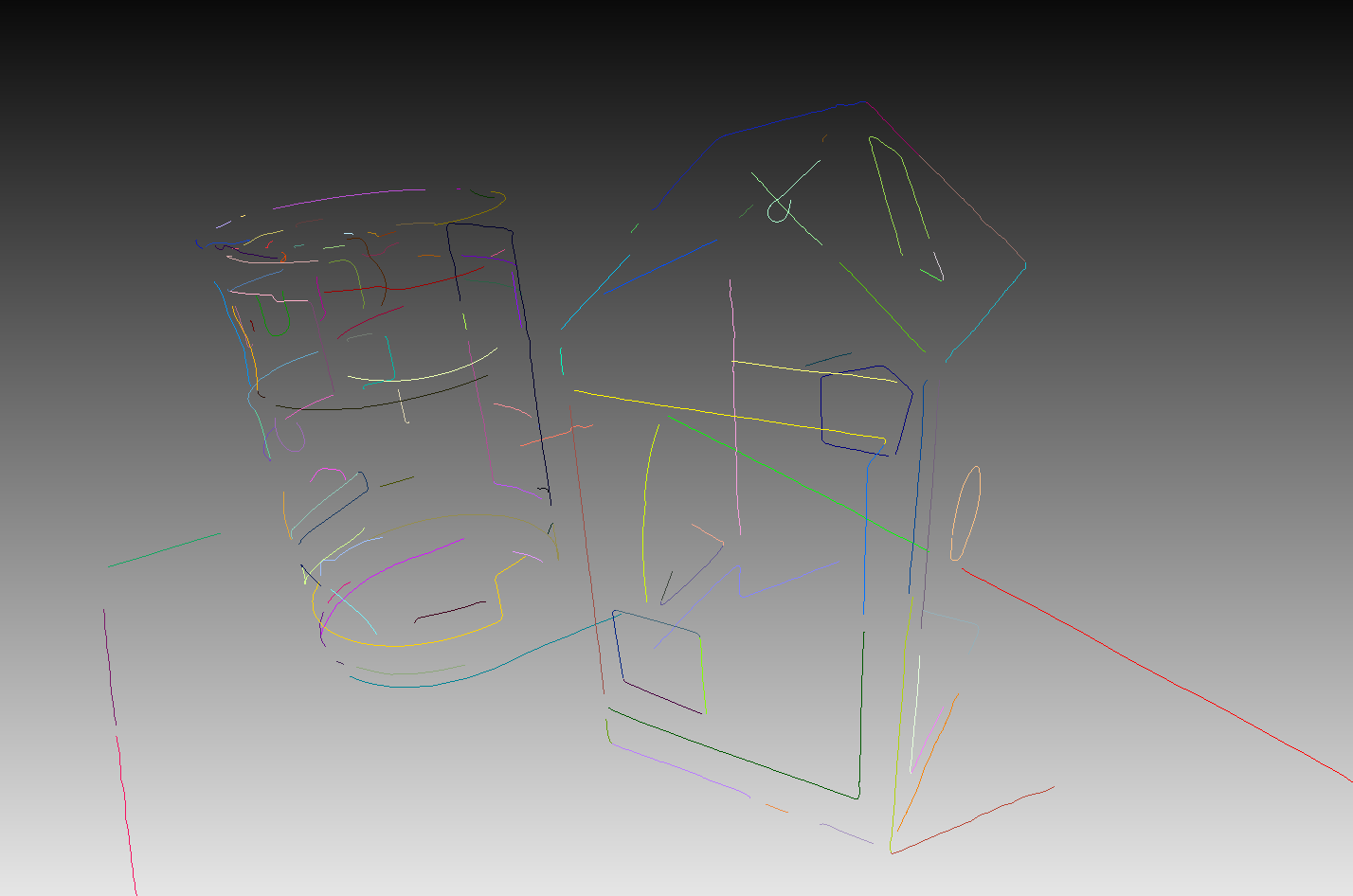}
    \includegraphics[width=0.49\linewidth]{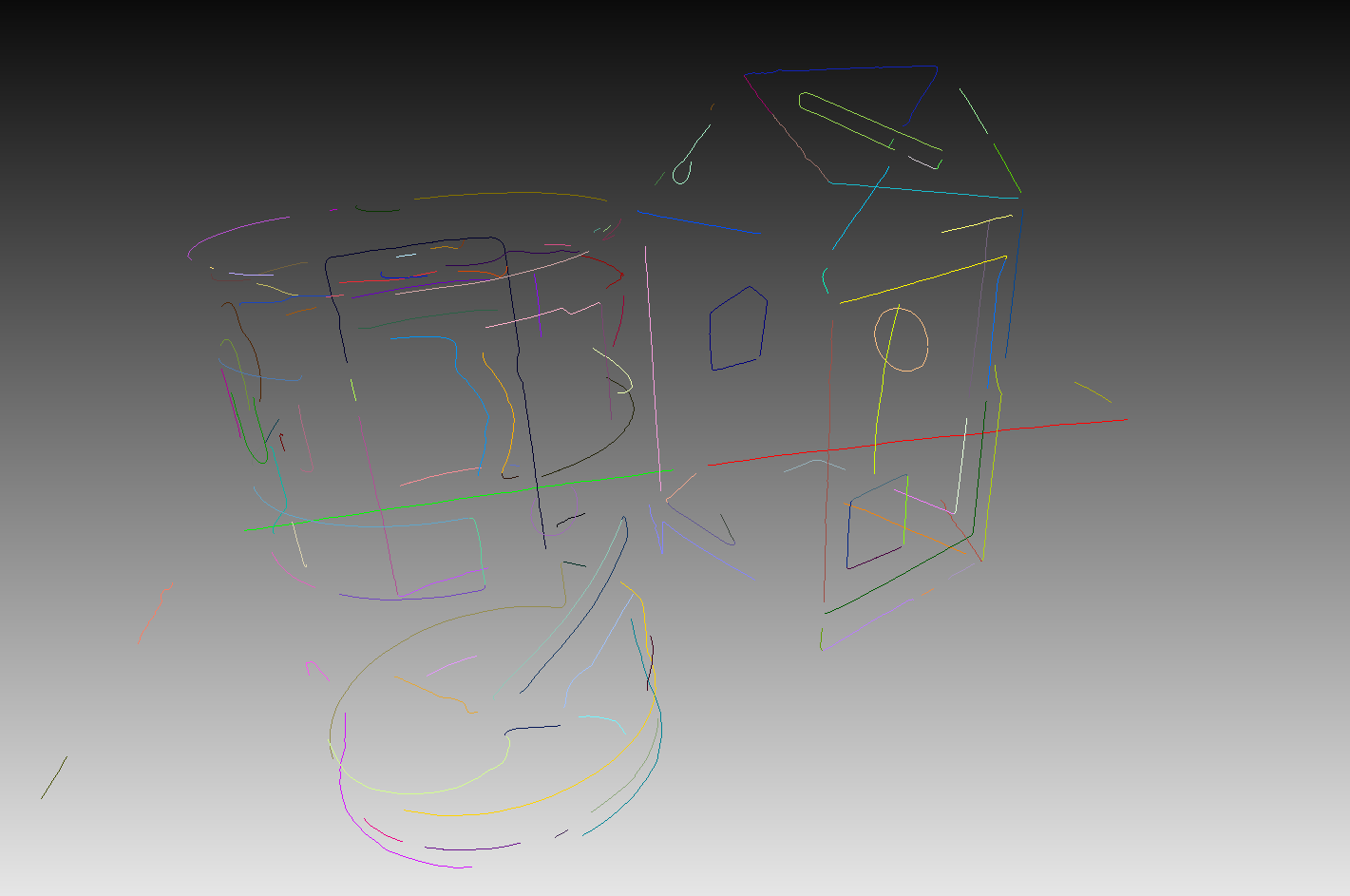}
  \end{center}
  \ReduceBeforeCaptionfigspace
  \caption{The original input 3d curve drawing (top row), which is the direct
  output of the 3D curve drawing approach, the result of our reorganization
  algorithm before breaking sharp corners (middle row), and after the sharp
  corners are broken (bottom row). The level of granularity displayed in the
  last row is the most appropriate for our lofting approach, as most surfaces
  are bounded by entire curves rather than subsegments.
  }
  \ReduceAfterCaptionfigspace
  \label{fig:reorg:results}
\end{figure}

In summary, this regrouping algorithm exploits the underlying organization, as
well as the rich differential geometric properties embedded in any
sufficiently-smooth, 3D curve representation, to adjust the granularity and
connectivity of any input curve graph or network to suit the needs of a wide
variety of applications. In the case of surface lofting, the quality of the
resulting reconstructions are significantly improved if the input curves that
have 3D surfaces between them have their samples more or less linearly aligned
with each other, resulting in a more robust quadrangulation step that kickstarts
most lofting approaches. We therefore use the 1st and 2nd order differential
geometric cues, namely tangents and curvatures, to full extent in order to
aggresively group smooth segments and break curves at high-curvature points,
maximizing the likelihood that the lofting algorithm will receive a set of 3D
curves best suited for its capabilities.

\section{Experiments and Results} \label{sec:lofting-results}
\begin{figure*}
  \begin{center}
    \includegraphics[width=0.245\linewidth]{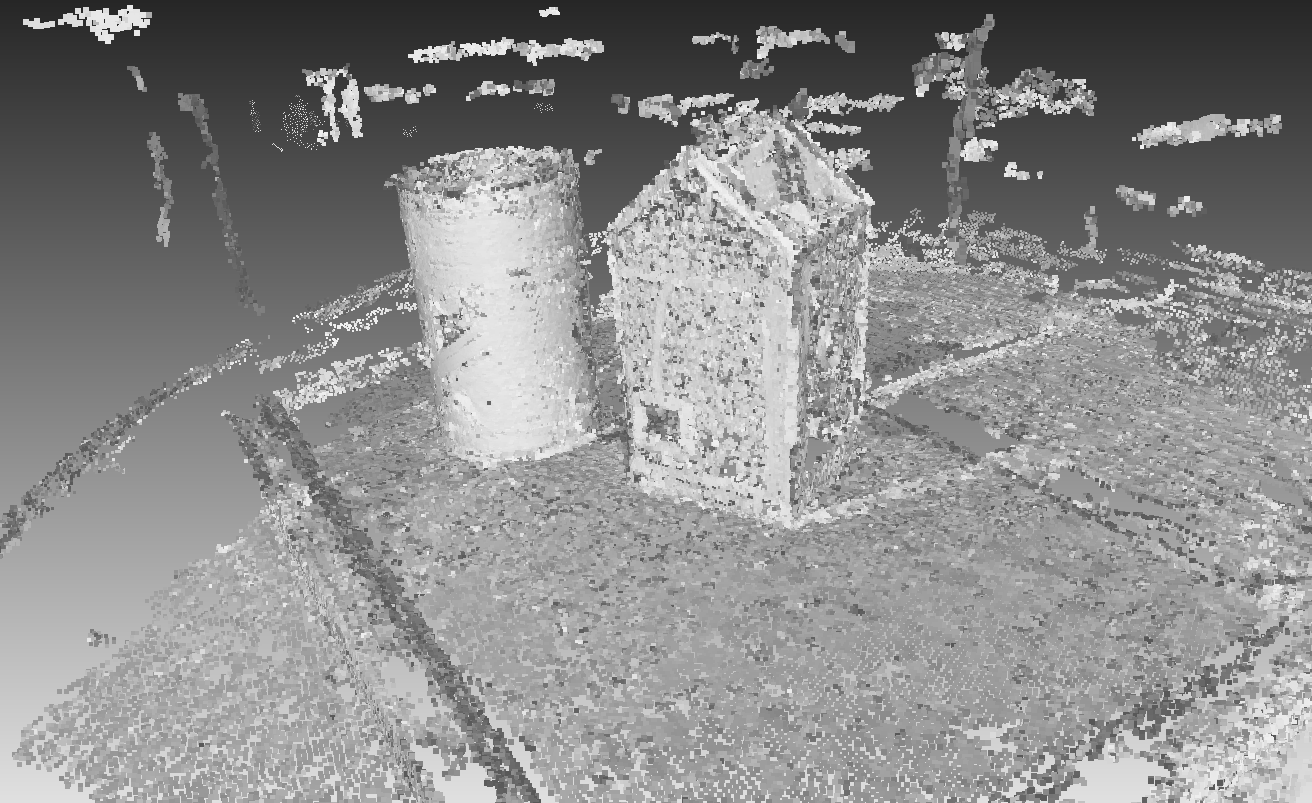}
    \includegraphics[width=0.245\linewidth]{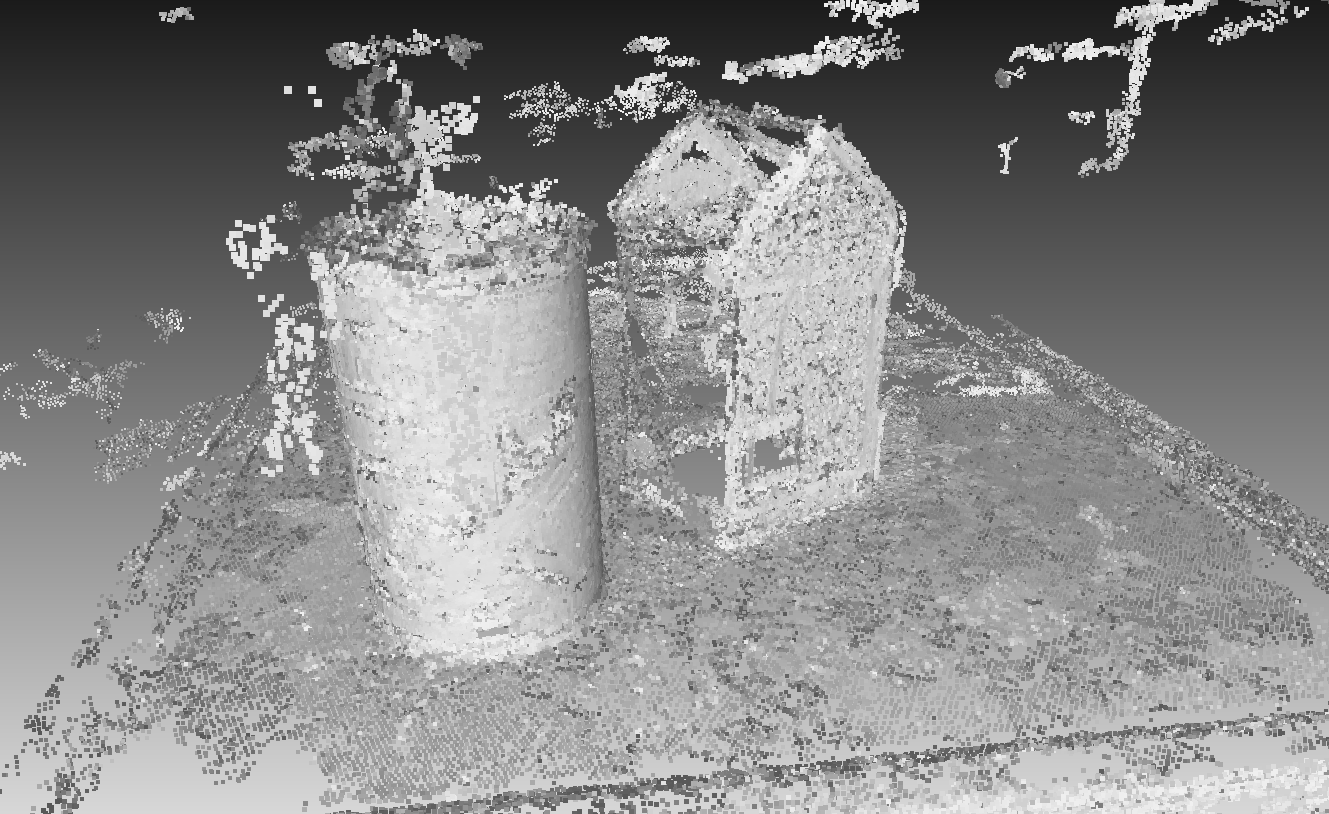}
    \includegraphics[width=0.245\linewidth]{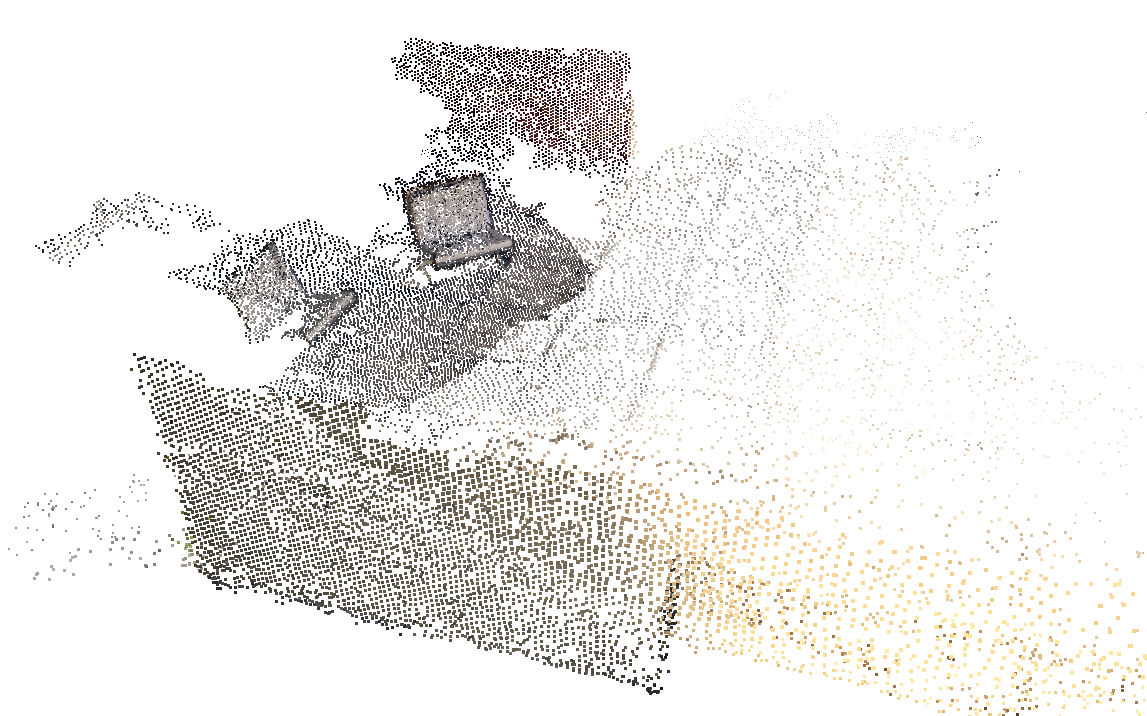}
    \includegraphics[width=0.245\linewidth]{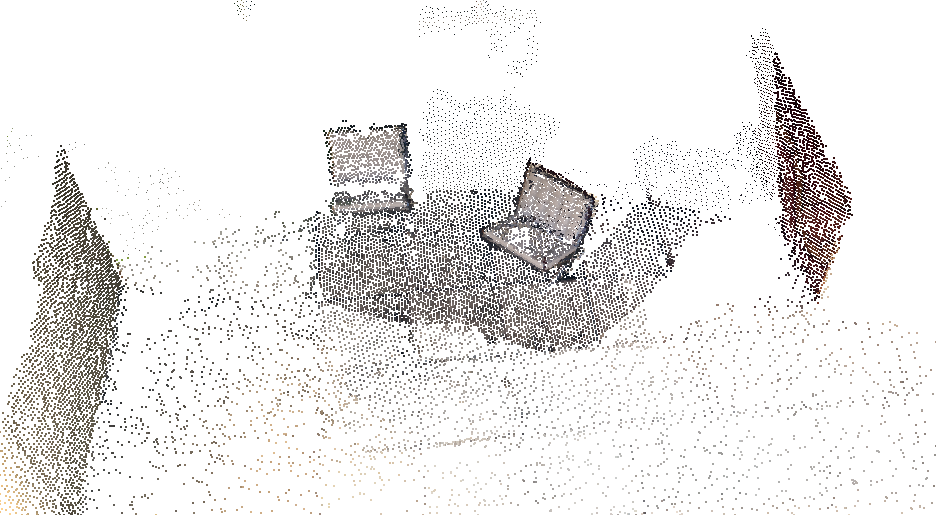}\\
    \includegraphics[width=0.245\linewidth]{figs/loftsurface_amsterdam_house_01.png}
    \includegraphics[width=0.245\linewidth]{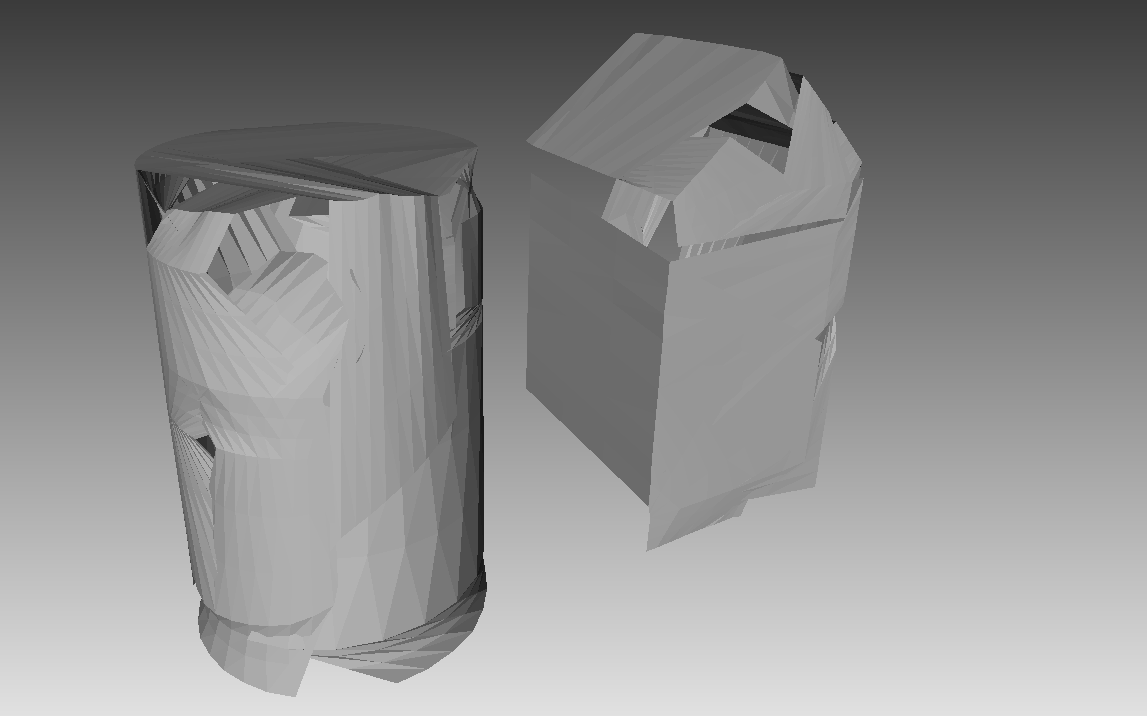}
    \includegraphics[width=0.245\linewidth]{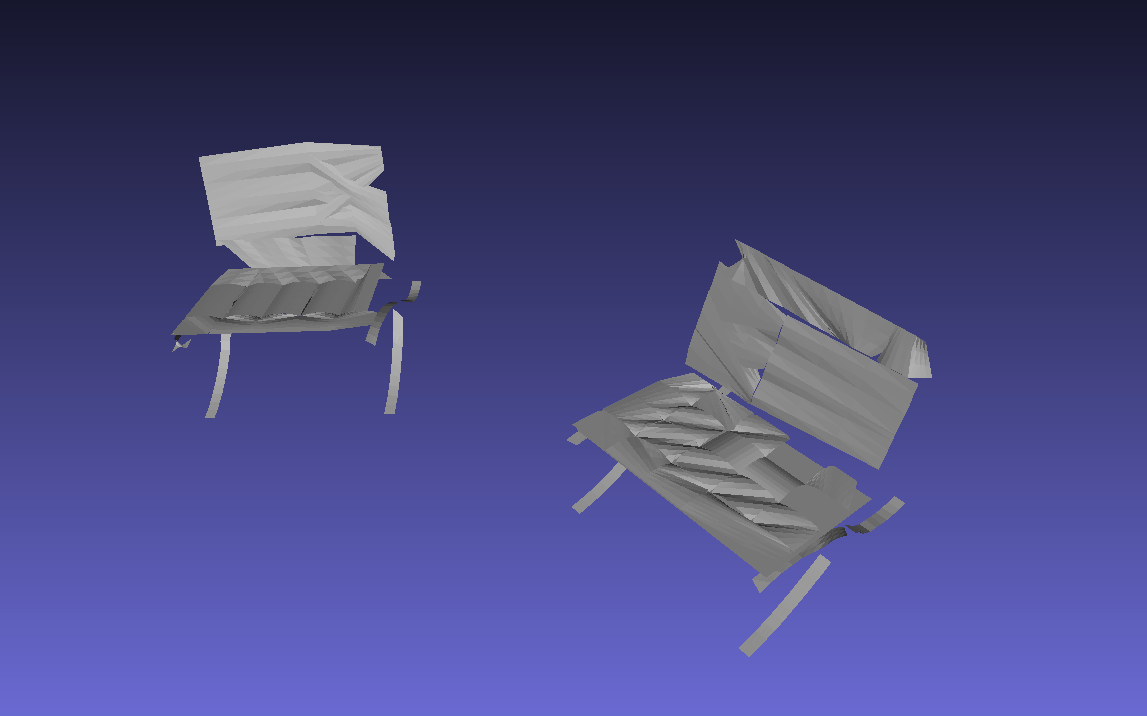}
    \includegraphics[width=0.245\linewidth]{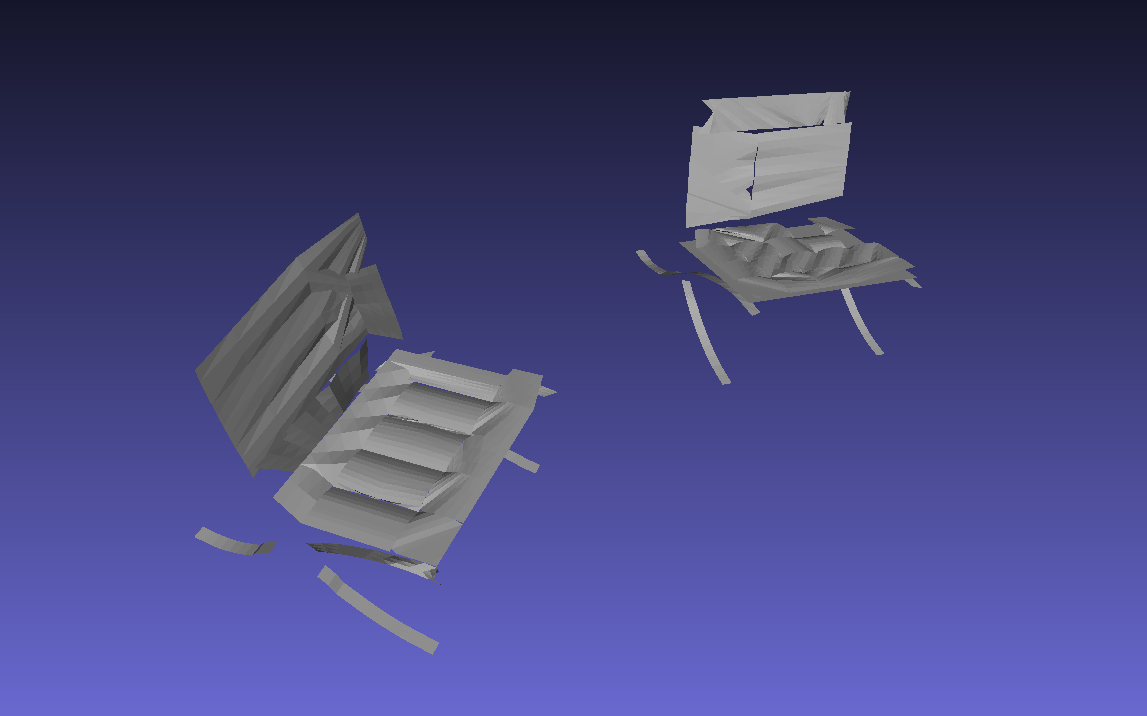}
  \end{center}
  \ReduceBeforeCaptionfigspace
  \caption{\small Two views of the PMVS reconstruction results on the Amsterdam
    House Dataset and Barcelona Pavilion Dataset (first row). Observe the wide
    gaps on homogeneous surfaces. The second row shows the results of our
    algorithm from the same views, obtained from a set of mere 27 curve
    fragments and without using appearance. Note that the PMVS gaps are filled
    in our results. Our algorithm errs in reconstructing the back of the can as
    a flat surface. This can easily be corrected via integration of appearance
    cues in the reconstruction process. 
  }\vspace{-0.3cm}
  \label{fig:lofting:results}
\end{figure*}

\noindent{\textbf{\underline{Implementation:}}} The 3D drawing is computed using
code made available by the authors of \cite{Usumezbas:Fabbri:Kimia:ECCV16}.
Smoothing code was made available by \cite{Garcia:CSDA2010}. We have selected
one of the most robust lofting implementations, BSurfaces, a part of
Blender~\cite{BSurfaces}, a well-known, professional-grade CAD system in
widespread use. BSurfaces is able to work on multiple curves with arbitrary
topology and configurations, either simultaneously or incrementally, producing
simple and smooth surfaces that accurately interpolate input curves, even if
they only partially cover the boundary of the surface to be reconstructed. The
use of BSurfaces has been limited to interactive modeling, where a human agent
provides clean well-connected curves to the system. 
To the best of our knowledge, a fully-automated 3D modeling pipeline that
obtains a 3D curve network, and uses lofting to surface this network in a
fully-automated fashion, is novel. 

\noindent{\textbf{\underline{Datasets:}}} We use two datasets to quantify
experimental results. First, the Amsterdam House Dataset consists of 50 fully
calibrated multiview images and comprises a wide variety of object properties,
including but not limited to smooth surfaces, shiny surfaces, specific
close-curve geometries, text, texture, clutter and cast shadows. This dataset is
used to evaluate the occlusion and visibility reasoning part of our pipeline,
Section~\ref{sec:auto-lofting}. Second, the Barcelona Pavilion Dataset is a
realistic synthetic dataset created for validating the present approach with
complete control over illumination, 3D geometry and cameras. This dataset was
used with its 3D mesh ground truth to evaluate the geometric accuracy of the
full pipeline.

\noindent{\textbf{\underline{Qualitative Evaluation:}}} Figure~\ref{fig:lofting:results} shows our algorithm's reconstruction and compares it to PMVS~\cite{Furukawa:Ponce:CVPR2007}. Observe that the reconstructed surface patches are glued onto the 3D drawing so that the topological relationship among surface patches is explicitly captured and represented. A key point to keep in mind is that the two approaches are not compared to see which is better. Rather, the intent is to show the complementary nature of the two appraches and the promise of even greater performance when appearance, the backbone of PMVS, is integrated into our approach.

\noindent{\textbf{\underline{Quantitative Evaluation:}}} The algorithm is quantitatively evaluated in two ways. First, we assume the input to the algorithm, the 3D curve drawing, is correct and compare ground truth to the algorithm's results based on a common 3D drawing. Specifically, we manually construct a surface model using the curve drawing in an interactive design and modeling context using Blender. The resulting surface model then serves as ground truth (GT) since it is the best possible expected outcome of our algorithm. Both GT and algorithm surface models are sampled and a proximity threshold is used to determine if a sample belongs to the other and vice versa. Three stages of surface reconstruction are then evaluated as a precision-recall curve, Figure~\ref{fig:lofting:quan}a, namely: {\em (i)} All surface hypotheses satisfying formation constraints; {\em (ii)} surface hypotheses that survive the occlusion constraint; {\em (iii)} surface hypotheses that further satisfy the visibility constraint with duplications removed. The algorithm recovers 90\% of the surfaces with nearly 100\% precision. The missing surfaces are those that do not occlude any structures, and therefore cannot be validated with our approach. Clearly, the use of appearance would a long way towards recovering these missing surfaces.

Second, we also quantitatively evaluate the algorithm in an end-to-end fashion,
including the 3D drawing stage. Since the ground truth surfaces are not
available from Amsterdam House Dataset, we resort to using Barcelona Pavilion
Dataset, which has GT surfaces. Since this dataset is large, we focus our
evaluation on a specific area with two chair objects. We use the same strategy
to compare the final outcome of our algorithm, Figure~\ref{fig:lofting:quan}b.
The results show that despite a complete disregard for appearance, geometry of
the surfaces together with occlusion constraint is able to recover a significant
number of surface patches accurately. The recall does not reach 100\% because
the ground truth floor surfaces do not occlude any curves and therefore cannot
be recovered.


\begin{figure}
  \begin{center}
    \includegraphics[width=\linewidth]{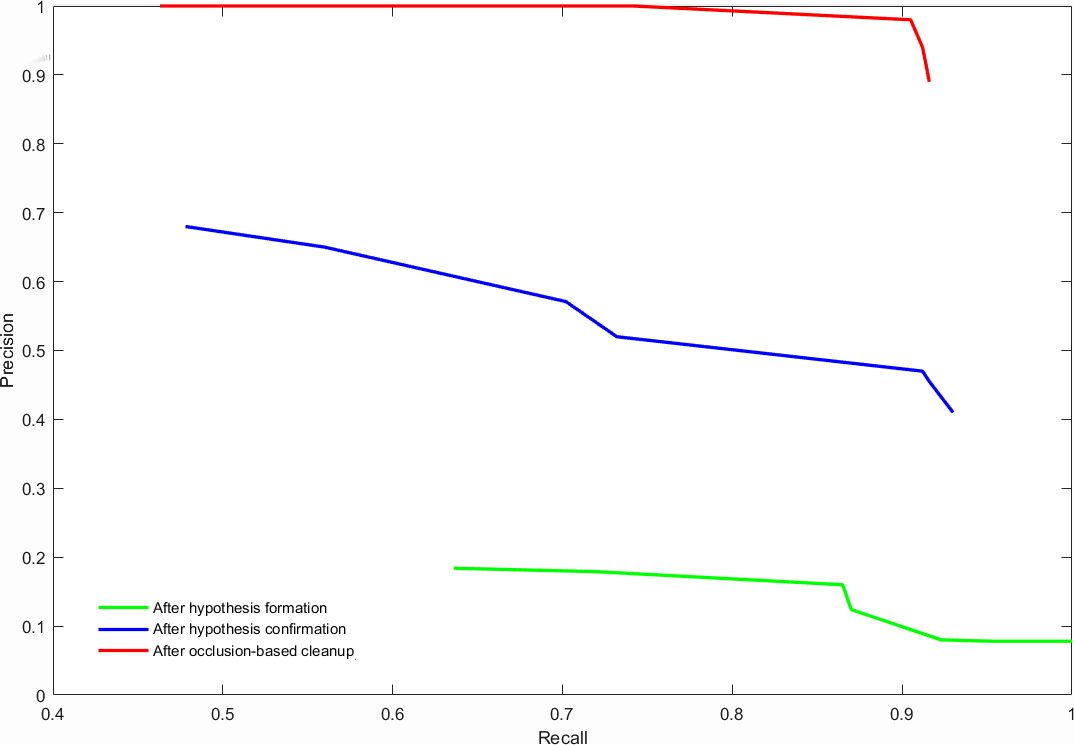}\\
    \includegraphics[width=\linewidth]{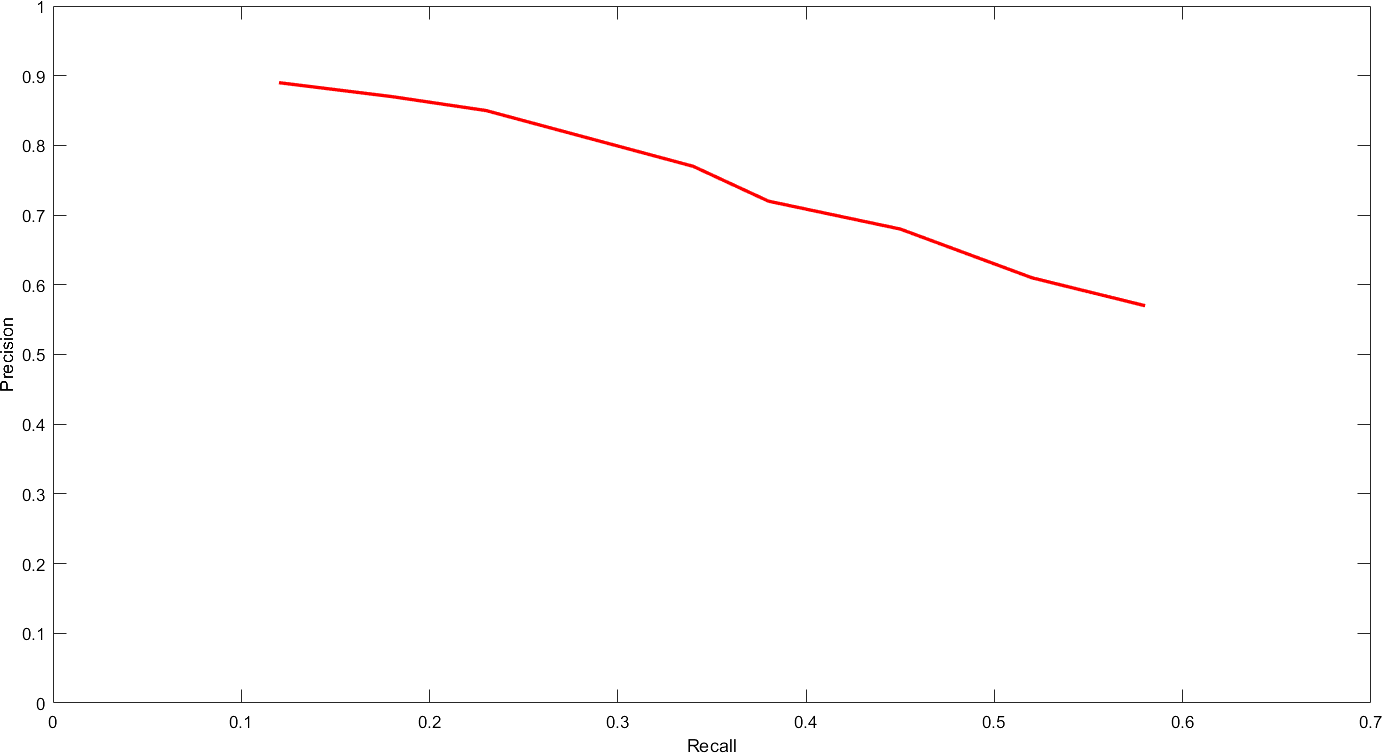}
  \end{center}
  \ReduceBeforeCaptionfigspace
  \caption{(a) The precision-recall curves for Amsterdam House Dataset,
    corresponding to post hypothesis-formation surfaces (green), confirmed
    surface (blue), and confirmed surfaces after occlusion-based cleanup (red).
    These results provide quantitative proof for the necessity of all steps in
    our reconstruction algorithm; (b) The precision-recall curve for Barcelona
    Pavilion Dataset, evaluating the geometric accuracy of the entire pipeline.
  }
  \label{fig:lofting:quan}
\end{figure}

\section{Conclusions}

This paper presents a fully automated dense surface reconstruction approach using geometry of curvilinear structure evident in wide baseline calibrated views of a scene. The algorithm relies on the {\em 3D drawing}, a graph-based representation of reconstructed 3D curve fragments which annotate meaningful structure in the scene, and on lofting to create surface patch hypotheses which are glued onto the 3D drawing, viewed as a scaffold of the scene. The algorithm validates these hypotheses by reasoning about occlusion among curves and surfaces. Thus it requires views from a wide range of camera angles and performs best if there are multiple objects to afford the opportunity for inter-object and intra-object occlusion. Qualitative and quantitative evaluations shows that a significant portion of the scene surface structure can be recovered and its topological structure is made explicit, a clear advantage. 
This is significant considering this is only the first step in our approach, namely, using geometry without using appearance which is the core idea underlying successful dense reconstruction systems like PMVS. Our goal is to integrate the use of appearance in the process which promises to significantly improve the reconstruction performance. 

{\small

}

\end{document}